\documentclass[pdflatex,sn-nature]{sn-jnl}

\usepackage{graphicx}%
\usepackage{float}%
\usepackage{amsmath,amssymb,amsfonts,bm}%
\usepackage{amsthm}%
\usepackage{bm}%
\usepackage{dsfont}
\usepackage{mathrsfs}%
\usepackage[title]{appendix}%
\usepackage{xcolor}%
\usepackage{textcomp}%
\usepackage{manyfoot}%
\usepackage{booktabs}%
\usepackage{algorithm}%
\usepackage{algorithmicx}%
\usepackage{algpseudocode}%
\usepackage{listings}%
\usepackage{enumitem}
\usepackage{verbatim}
\usepackage{epsfig}
\usepackage{epstopdf}
\usepackage{helvet}
\usepackage{courier}
\usepackage{upgreek}
\usepackage[bottom]{footmisc}
\usepackage{longtable}

\usepackage{geometry}            
\newtheorem*{theorem*}{Theorem}

\makeatletter
\def\paragraphfont{\reset@font\fontfamily{\rmdefault}\fontsize{10bp}{12bp}\bfseries\selectfont\raggedright}

\renewcommand\paragraph{\@startsection{paragraph}{4}{\z@}%
                                      {-12pt \@plus -4pt \@minus-2pt}
                                      {3pt}
                                      {\paragraphfont}}
\makeatother

\graphicspath{{picture/}} 

\begin{document}

\title[End-to-End OAL Stability]{End-to-End Stability of Online Active Learning in Large-Scale Neural Systems}

\author[1,7,*]{Shuiwen Shen}
\author[2]{Jinhua Zhang}
\author[3]{Chenlei Hou}
\author[4]{Peng Shi}
\author[1]{Yong Han}
\author[5]{\protect\\ Bo Gao}
\author[6]{Xiaoxiao Na}
\author[1]{Zhihui Lin}
\author[1]{Pengtao Du}

\affil[1]{School of Mechanical and Automotive Engineering, Xiamen University of Technology, Xiamen, 361024, China}
\affil[2]{China Innovation Center for Intelligent and Connected Vehicles, Beijing, 100176, China}
\affil[3]{King Long United Automotive Industry Co. Ltd., Xiamen, Fujian, 361023, China}
\affil[4]{School of Electrical and Mechanical Engineering, Adelaide University,  Adelaide, SA 5005, Australia}
\affil[5]{Syselek UK Ltd, Warwick, CV34 6RG, United Kingdom}
\affil[6]{Department of Engineering, Cambridge University, Cambridge, CB2 1TN, United Kingdom}
\affil[7]{State Innovative Centre for Automotive Electric Driving at Xiamen, Xiamen, Fujian, 361010, China}
\affil[*]{2023000085@xmut.edu.cn}

\abstract {\hyphenpenalty=10000\exhyphenpenalty=10000 Online Active Learning (\textbf{OAL}) aims to extend deep learning from static inference after offline training to continuous adaptation during real-time interaction with a changing physical environment. This transition is essential for safety critical applications, but it raises a central theoretical question: whether the online parameter update process remains stable when scaled to multilayer neural architectures. This paper develops a stability focused \textbf{OAL} framework. The framework uses prediction error as a feedback signal to update a structured deep model, thereby linking online learning with Lyapunov based energy minimization. We show that prediction error convergence alone is insufficient for global parameter convergence because multilayer networks may admit high dimensional manifolds of equivalent output solutions. The analysis therefore separates output error regulation, internal zero dynamics, interlayer coupling, and parameter convergence. Persistency of excitation and active damping are then used to remove the residual null space modes that remain after output error convergence. A cascaded Lyapunov analysis establishes asymptotic stability of the overall \textbf{OAL} learning process, while monotonic activation functions and diffeomorphic layer mappings provide the uniqueness condition for the target parameter trajectory. The resulting proof provides a theoretical basis for applying online adaptive learning to large scale neural systems in nonstationary and safety critical settings.
}


\maketitle

\thispagestyle{empty}
\section{Introduction}

Online learning is essential for applications operating in non-stationary physical environments. In contrast to offline-trained deep learning models, whose parameters are typically fixed after deployment, an online learning system must continuously update its internal model according to real-time sensory feedback. This capability is particularly important for safety-critical systems, where distributional shifts, unmodeled disturbances, and out-of-distribution inputs can cause prediction errors to accumulate during operation. However, continuous parameter adaptation also introduces a fundamental theoretical challenge: the learning process itself must remain stable.

Existing deep learning architectures are highly scalable, but their stability is usually established within an offline optimization setting. Backpropagation provides an effective mechanism for minimizing empirical loss over a fixed dataset, yet it does not directly guarantee stable real-time adaptation when data arrive sequentially from a changing physical environment \cite{Rumelhart1986BackProp,LeCun2015DeepLearning}. Concerns about global credit assignment and biological plausibility have consequently motivated more local learning alternatives, including the Forward--Forward algorithm \cite{hinton2022forward}. More broadly, biological intelligence relies on inherited structure and recurrent organization that are not captured by pure learning alone \cite{zador2019critique}. Continual-learning and test-time-adaptation methods likewise seek to update deployed models under distribution shift \cite{delange2021continual,parisi2019continual,wang2020tent}, but their optimization objectives do not by themselves provide an end-to-end Lyapunov stability guarantee. Classical adaptive learning and control theory, by contrast, provide mature tools for proving convergence and stability through Lyapunov analysis, but these methods are often limited by the difficulty of representing large-scale nonlinear systems with high-dimensional parameter spaces. This creates a gap between the scalability of deep learning and the formal stability requirements of online adaptive systems.

Online Active Learning (\textbf{OAL}) is introduced to bridge this gap. The central idea of \textbf{OAL} is to use prediction error as a continuous feedback signal for updating a structured deep model in real time. Rather than treating learning as a separate offline training phase, \textbf{OAL} embeds learning into the system's closed-loop interaction with the external world. The predictive model receives sensory input, generates an internal prediction, evaluates the discrepancy between prediction and observation, and updates its parameters through an error-feedback learning law. In this sense, \textbf{OAL} transforms deep learning from a static inference architecture into a dynamic adaptive system.

The key question addressed in this paper is whether such an online learning process can be made stable at large scale. A naive error-feedback update can reduce the output prediction error, but it does not automatically guarantee convergence of the internal parameters. In multilayer neural systems, different parameter configurations may produce the same output, leaving the learning trajectory on a high-dimensional manifold of equivalent solutions. As a result, prediction-error convergence alone is insufficient for proving global learning stability. A stability proof must therefore account not only for the decay of output error, but also for parameter convergence, cross-layer coupling, and internal zero dynamics.

This paper develops a stability-centered analysis of \textbf{OAL}. Section~\ref{sec_OAL} formulates the primary \textbf{OAL} framework and its error-feedback learning law. Section~\ref{sec_stability} then analyzes why non-increasing prediction-error energy does not by itself imply parameter convergence. The proof separates the marginal zero dynamics from the damped cascaded dynamics, identifies the excitation and damping conditions needed to remove the null-space ambiguity, and derives the resulting global Lyapunov stability condition.

\section{Principle of Online Active Learning}
\label{sec_OAL}

\subsection{Deep learning and adaptive learning foundations}

Deep learning and adaptive learning provide complementary foundations for online model adaptation. Deep learning offers a scalable representation mechanism for high-dimensional nonlinear mappings, while adaptive learning provides a feedback-based view of how a model can change during operation. In a conventional deep neural network, the predictive output can be written abstractly as
\begin{equation}
  \hat{\mathbf{y}}={f\mathbb{NN}}_\mathrm{dl}(\mathbf{x},\bm{\uptheta}),
  \label{eqn_gDL}
\end{equation}
where \(\mathbf{x}\) denotes the input, \(\hat{\mathbf{y}}\) the predicted output, and \(\bm{\uptheta}\) the trainable parameters. The corresponding layer-wise architecture is denoted by
\begin{equation}
  \mathbf{z}^{[i]} = \bm{\uptheta}^{[i]}\mathbf{y}^{[i-1]}, \qquad
  \mathbf{y}^{[i]} = g_i(\mathbf{z}^{[i]}), \qquad i=1,2,\ldots,n,
  \label{eqn_DLArch}
\end{equation}
where \(g_i\) is the nonlinear activation function of layer \(i\). This linear-layer-plus-activation structure gives deep learning its representational capacity \cite{zhou2020Representation} and preserves a repeated layer-wise form across the network. Successive layers can also be interpreted as progressively transforming or compressing the information needed for the downstream task \cite{amjad2019learning}. However, in the standard offline paradigm, the parameters are optimized before deployment and are then held fixed during inference.

The offline optimization process is commonly driven by backpropagation. If \(\mathcal{L}\) denotes the training loss, the layer-wise gradient recursion can be summarized as
\begin{equation}
  \begin{array}{rcl}
    \mathbf{g}_{i} &=& \dfrac{\partial \mathbf{y}^{[i+1]}}{\partial \mathbf{y}^{[i]}}\mathbf{g}_{i+1},\\[2mm]
    \mathbf{g}_{\bm{\uptheta}_i} &=& \dfrac{\partial \mathbf{y}^{[i]}}{\partial \bm{\uptheta}^{[i]}}\mathbf{g}_{i},\\[2mm]
    \Delta\bm{\uptheta}^{[i]} &=& -\lambda \mathbf{g}_{\bm{\uptheta}_i}.
  \end{array}
  \label{eqn_DL_BP}
\end{equation}
Here \(\mathbf{g}_{i}\) and \(\mathbf{g}_{\bm{\uptheta}_i}\) are the gradients associated with \(\mathbf{y}^{[i]}\) and \(\bm{\uptheta}^{[i]}\), respectively. This recursive structure is one reason deep networks can remain trainable at very large parameter scales. It also provides the structural template later used to interpret online feedback learning as a layer-wise isomorphic process. However, this mechanism is designed for learning from a fixed dataset and does not by itself define a stable online adaptive system. Once the external environment changes, the model may accumulate prediction error without any principled way to update its parameters during operation.

Adaptive learning starts from the opposite perspective. Rooted in adaptive-control theory \cite{aastrom1995adaptive,ioannou1996robust}, it represents the process by a parameterized state-space model and treats prediction error as a real-time feedback signal. For an adaptive model with estimated state \(\hat{\mathbf{s}}\), input \(\mathbf{x}\), and predicted output \(\hat{\mathbf{y}}\), the model can be described abstractly as
\begin{equation}
  \dot{\hat{\mathbf{s}}}=f(\hat{\mathbf{s}},\bm{\uptheta},\mathbf{x}), \qquad
  \hat{\mathbf{y}}=g(\hat{\mathbf{s}},\bm{\uptheta}).
  \label{eqn_APStateFun}
\end{equation}
The tracking error is then defined as
\begin{equation}
  \mathbf{e}_y = \mathbf{y}_s-\hat{\mathbf{y}}_s,
  \label{eqn_APErrFun}
\end{equation}
where \(\mathbf{y}_s\) and \(\hat{\mathbf{y}}_s\) are the measured and predicted sensory outputs, respectively. Adaptive learning updates the model according to this feedback. In a generic form, the learning dynamics can be written as
\begin{equation}
  \dot{\bm{\uptheta}} = \lambda h(\mathbf{e}_y,\hat{\mathbf{s}}),
  \label{eqn_APLearn_simple}
\end{equation}
where \(\lambda\) is an adaptation gain. If the mapping from \(\bm{\uptheta}\cdot\hat{\mathbf{s}}\) to \(\mathbf{e}_y\) is Strictly Positive Real (SPR), the following simplified learning law can guarantee stability and solution uniqueness \cite{slotine1991applied}:
\begin{equation}
  \dot{\bm{\uptheta}} = \lambda \mathbf{e}_y\hat{\mathbf{s}}.
  \label{eqn_APLearn}
\end{equation}
This feedback view is attractive for online operation, but the SPR condition is often violated in large nonlinear systems, particularly in the presence of unmodeled dynamics. Moreover, the solution to \(\mathbf{e}_y=0\) is not necessarily unique in high-dimensional parameter spaces. Classical adaptive formulations are therefore difficult to scale directly to deep models with many layers and many parameters \cite{aastrom1995adaptive,ioannou1996robust}.

\subsection{Primary OAL framework and error-feedback learning law}
\label{Sec_OAL_Frm}

\begin{figure}[H]
  \centering
  \begin{minipage}[b]{0.49\linewidth}
    \centering
    \includegraphics[width=\linewidth,height=0.20\textheight]{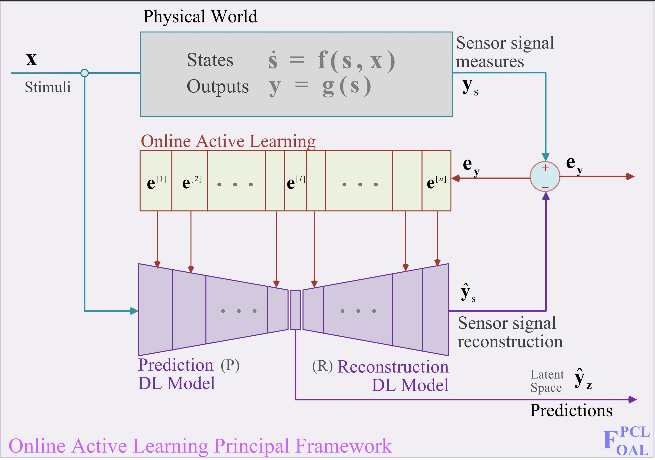}\\[-0.5ex]
    \textbf{(a)} Principal framework
  \end{minipage}\hfill
  \begin{minipage}[b]{0.49\linewidth}
    \centering
    \includegraphics[width=\linewidth,height=0.20\textheight]{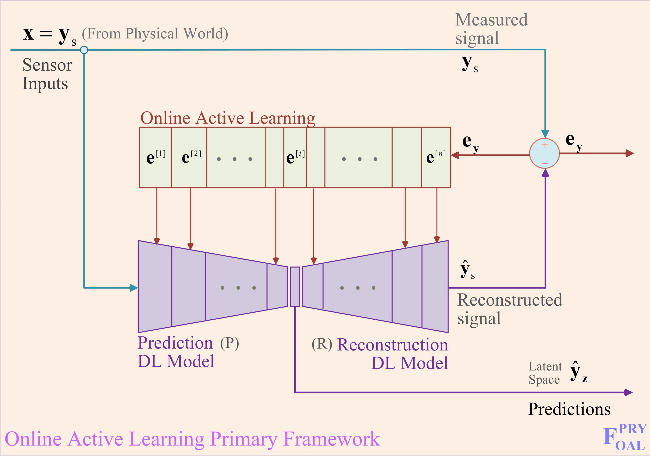}\\[-0.5ex]
    \textbf{(b)} Primary framework
  \end{minipage}
  \caption{From the principal Online Active Learning framework to the primary framework. \textbf{(a)} The external stimulus $\mathbf{x}$ drives both the true physical process and the structured deep model, while the sensor measurement $\mathbf{y}$ is the comparison target. The \emph{Physical World} block denotes the true process, whereas Eqn~(\ref{eqn_APStateFun}) describes its parameterized internal model. The gradient arrows correspond to $\mathbf{J}_\uptheta^{[i]}\mathbf{e}^{[i]}$ in Eqn~(\ref{eqn_OAL_Update}), up to the gradient-sign convention. \textbf{(b)} The primary framework $\mathbb{F}_\mathrm{OAL}^\mathrm{PRY}$ sets $\mathbf{x}=\mathbf{y}_s$ and decomposes the model into prediction $\mathbb{P}$ and reconstruction $\mathbb{R}$, with the resulting error $\mathbf{e}_y$ distributed across the layers for online parameter correction.}
  \label{fig:OALPrimary}
\end{figure}

Online Active Learning (\textbf{OAL}) combines the representational capacity of deep learning with the feedback structure of adaptive learning. In its conceptual form, a single deep model as in Eqn~(\ref{eqn_gDL}) is corrected online by the sensory prediction error, as illustrated in Fig.~\ref{fig:OALPrimary}a, where the model is driven by an external stimulus $\mathbf{x}$ while the sensor measurement $\mathbf{y}$ serves as the comparison target. This view keeps the layer-wise error propagation structure of backpropagation \cite{Rumelhart1986BackProp,LeCun2015DeepLearning,lillicrap2020backpropagation}, but replaces the offline loss gradient with an online error signal generated by the mismatch between physical-world measurements and model predictions. It is also consistent with the hierarchical error-propagation idea in predictive coding \cite{rao1999predictive,friston2009theory,Hosoya2005PC,Huang2011PC} and with the Bayesian view that perception continuously reconciles internal predictions with uncertain sensory evidence \cite{knill2004bayesian}. Related active-inference models similarly couple prediction and learning through generative state-space dynamics \cite{ccatal2020learning}. Unlike the cited formulations, which emphasize inference over latent states, the present analysis focuses on Lyapunov stability of the parameter-adaptation dynamics in a large-scale layered model.

The conceptual framework is already sufficient to drive the predicted output towards the measurement, but it leaves two questions open. The error-driven correction alone provides no mechanism for extracting a compact internal representation of the external world, and it gives no guarantee that the model parameters themselves converge. From a system-dynamics perspective the second question is the more fundamental one, because the same output can be produced by many different parameter configurations. The first question is answered by restructuring the model, as described below; the second is the subject of Section~\ref{sec_stability}.

To obtain a representation, the monolithic model of Eqn~(\ref{eqn_gDL}) is decomposed into a prediction model $\mathbb{P}$, which compresses the sensory input into a low-dimensional latent space, an optional latent model $\mathbb{L}$, which evolves the latent state, and a reconstruction model $\mathbb{R}$, which maps the latent state back to the observation space. Predictive tasks are then carried out in the latent space, while the reconstruction supplies the learning signal. Setting the model input directly to the sensory stream, $\mathbf{x}=\mathbf{y}_s$, yields the primary framework $\mathbb{F}_\mathrm{OAL}^\mathrm{PRY}$ shown in Fig.~\ref{fig:OALPrimary}b. The system then interacts with the external world exclusively through its sensory boundary, in the manner of a Markov blanket \cite{kirchhoff2018markov}, and can learn from a continuously arriving stream without an externally supplied label.

The framework operates as a repeated prediction--evaluation--correction cycle. During prediction, $\mathbf{y}_s$ passes through the prediction model $\mathbb{P}$ to form the latent representation $\hat{\mathbf{y}}_z$, after which the reconstruction model $\mathbb{R}$ produces $\hat{\mathbf{y}}_s$. Both models are layered neural networks of the form defined in Eqn~(\ref{eqn_DLArch}), and their complete forward mapping is represented by Eqn~(\ref{eqn_gDL}). In the primary diagram, any latent dynamics model $\mathbb{L}$ is implicit and may be absorbed into $\mathbb{P}$ or $\mathbb{R}$. It is written explicitly later only to express the general learning dynamics used in the stability analysis. The latent representation $\hat{\mathbf{y}}_z$ is also the framework output marked as \emph{Predictions} in Fig.~\ref{fig:OALPrimary}b. It is consumed by the downstream task, whereas reconstruction is used to generate the online learning signal. For example, in a visual perception system, $\mathbb{P}$ may compress the incoming camera or sensor stream into a latent description of lane geometry, obstacles, or motion context. The reconstruction branch then tests whether this latent state still explains the current sensory signal. The reconstruction error provides the online correction signal, while the latent prediction remains available for planning or control.

During evaluation, the reconstructed signal is compared with the current sensory measurement according to Eqn~(\ref{eqn_APErrFun}). During correction, this same error is propagated backward from $\mathbb{R}$ to $\mathbb{P}$ and updates every constituent layer. The corrected model then processes the next measurement, so inference and learning proceed concurrently. This closed loop provides the organizing structure for all equations below.

Because the reconstruction target is the input itself, the primary framework admits the degenerate solution $\mathbb{R}=\mathbb{P}^{-1}$, which drives the error to zero without capturing any physical structure of the environment. The practical spatial and temporal variants of \textbf{OAL} break this symmetry by reconstructing a different sensory channel or the next frame in time. Both retain the encoder--latent--decoder structure and the error-feedback learning law of Fig.~\ref{fig:OALPrimary}b, so the stability analysis below applies to them without modification.

The basic energy quantity minimized by \textbf{OAL} is the prediction-error energy
\begin{equation}
  \mathcal{L}_e = \frac{1}{2}\mathbf{e}_y^T\mathbf{e}_y,
  \label{eqn_Energy}
\end{equation}
where \(\mathbf{e}_y\) is the sensory prediction error generated at the comparison node in Fig.~\ref{fig:OALPrimary}b. Inside the \emph{Online Active Learning} block, this error is propagated backward and converted into $\mathbf{e}^{[1]},\ldots,\mathbf{e}^{[n]}$. Each layer receives its corresponding error signal for parameter adaptation. The local error of layer \(i\) is written as
\begin{equation}
  \mathbf{e}^{[i]} = \mathbf{J}_y^{[i,i+1]}\mathbf{e}^{[i+1]},
  \label{eqn_OAL_BP}
\end{equation}
where \(\mathbf{J}_y^{[i,i+1]}\) denotes the backward output sensitivity between adjacent layers. The corresponding parameter sensitivity is
\begin{equation}
  \mathbf{J}_\uptheta^{[i]}=\frac{\partial \mathbf{y}^{[i]}}{\partial\bm{\uptheta}^{[i]}},
  \label{eqn_OAL_JCBT}
\end{equation}
and the local online update law is summarized as
\begin{equation}
  \Delta\bm{\uptheta}^{[i]}=\lambda_i \mathbf{J}_\uptheta^{[i]}\mathbf{e}^{[i]}.
  \label{eqn_OAL_Update}
\end{equation}
For brevity, \(\mathbf{J}_y^{[i,i+1]}\) is written as \(\mathbf{J}_y^{[i]}\), with \(\mathbf{J}_y^{[n]}=\mathbf{I}\). The global sensitivities along the backward path are
\begin{equation}
  \mathbb{J}_y^{[i]} = \prod_{k=i}^{n}\mathbf{J}_y^{[k]}.
  \label{eqn_OverAll_JCBY}
\end{equation}
The corresponding global parameter sensitivity is
\begin{equation}
  \mathbb{J}_\uptheta^{[i]} = \mathbf{J}_\uptheta^{[i]}\mathbb{J}_y^{[i]},
  \label{eqn_OverAll_JCBT}
\end{equation}
so that the learning law can also be written in global form as
\begin{equation}
  \Delta\bm{\uptheta}^{[i]} = \lambda_i\mathbb{J}_\uptheta^{[i]}\mathbf{e}_y.
  \label{eqn_OAL_GL_Update}
\end{equation}
This global form shows how the single output error in Fig.~\ref{fig:OALPrimary}b reaches every layer of $\mathbb{R}$ and $\mathbb{P}$. The model changes online because each layer receives an error-driven correction, but the stability of the resulting coupled multilayer dynamics still has to be proven.

At the prediction-error level, the \textbf{OAL} update law leads to a Lyapunov-like descent condition. Under bounded external variation and positive learning gain, the derivative of the prediction-error energy can be expressed in the simplified form
\begin{equation}
  \dot{\mathcal{L}}_e = -\lambda\mathbf{e}_y^T\mathbb{J}_\uptheta^T\mathbb{J}_\uptheta\mathbf{e}_y \leq 0.
  \label{eqn_OAL_GAStb}
\end{equation}
A computationally relaxed update may also use the sign of the sensitivity matrix,
\begin{equation}
  \Delta\bm{\uptheta}^{[i]}=\lambda_i\,\mathrm{sgn}(\mathbb{J}_\uptheta^{[i]})\mathbf{e}^{[i]},
  \label{eqn_OAL_FlxUpdate}
\end{equation}
because \(\mathrm{sgn}(\mathbb{J}_\uptheta)^T\mathbb{J}_\uptheta\geq 0\) preserves the non-negative sensitivity product required by the stability argument while avoiding full Jacobian evaluation at every step.

For the stability proof, the latent model $\mathbb{L}$ that is implicit in Fig.~\ref{fig:OALPrimary}b is written out explicitly. Propagating the prediction error through \(\mathbb{R}\to\mathbb{L}\to\mathbb{P}\) gives the damped learning laws
\begin{align}
  \dot{\bm{\uptheta}}_R + \kappa_R \bm{\uptheta}_R &= \lambda_R \mathbf{J}_{\uptheta_R}^T \mathbf{e}_R, & \mathbf{e}_R &= \mathbf{e}_y, \label{eqn_OAL_LearnDyn_R}\\
  \dot{\bm{\uptheta}}_L + \kappa_L \bm{\uptheta}_L &= \lambda_L \mathbf{J}_{\uptheta_L} \mathbf{e}_L, & \mathbf{e}_L &= \mathbf{J}_L \mathbf{e}_R, \label{eqn_OAL_LearnDyn_L}\\
  \dot{\bm{\uptheta}}_P + \kappa_P \bm{\uptheta}_P &= \lambda_P \mathbf{J}_{\uptheta_P} \mathbf{e}_P, & \mathbf{e}_P &= \mathbf{J}_P \mathbf{e}_L, \label{eqn_OAL_LearnDyn_P}
\end{align}
where the learning inertia coefficient \(\kappa_{(\cdot)}>0\) originates from parameter regularization and acts as a damping term, written layer-wise as \(\mathbb{K}_\uptheta^{[i]}\) in Section~\ref{sec_stability}, that preserves internal priors and enhances real-time learning stability. These equations establish the principle that \textbf{OAL} learning consists of feedback-driven adaptive parameter loops with explicit damping. The following section therefore focuses on the central theoretical question: whether these coupled loops remain globally stable when extended to large-scale multilayer neural systems.
\section{Stability Proof of Online Active Learning}
\label{sec_stability}

\subsection{End-to-end global learning stability}

As shown in Subsection~\ref{Sec_OAL_Frm}, the \textbf{OAL} error-feedback law yields a non-increasing prediction-error energy under the stated bounded-variation and positive-gain assumptions. This property provides the basic online error-regulation mechanism for \textbf{AI} models. However, non-increasing output error energy alone does not guarantee the convergence of self-adapting parameters, which is addressed in this section.

\subsubsection*{Root of the problem with global convergence}

Consider the parameters $\bm{\uptheta}$ of the deep learning model defined in Eqn(\ref{eqn_gDL}), represented by the tensor
\[
\bm{\uptheta} = [\bm{\uptheta}^{[1]}, \bm{\uptheta}^{[2]}, \cdots, \bm{\uptheta}^{[i]}, \cdots \bm{\uptheta}^{[n]} ]^T
\]
where $\bm{\uptheta}^{[i]}$ denotes the parameters for the $i$-th layer of the deep neural network model spanning the prediction, latent, and reconstruction models $\mathbb{P}$, $\mathbb{L}$, and $\mathbb{R}$, respectively. This formulation generalizes to Convolutional Neural Networks (CNNs) and Transformer architectures, which can be mathematically treated as sparse variants of $\bm{\uptheta}^{[i]}$, in general. Matrix $\bm{\uptheta}^{[i]}$ is decomposed as
\begin{equation}
\bm{\uptheta}^{[i]} =
[\bm{\uptheta}_1^{[i]}, \bm{\uptheta}_2^{[i]}, \cdots, \bm{\uptheta}_k^{[i]}, \cdots \bm{\uptheta}_{n_i}^{[i]} ]^T
\label{eqn_Theta_Matrix}
\end{equation}
in which $n_i$ indicates the width of the corresponding layer, and $\bm{\uptheta}_k^{[i]}$ the weight vector for the  $k$-th node of $i$-th layer. Performing a first order Taylor expansion of the predictive error $\textbf{e}_y$ as in Eqn(\ref{eqn_APErrFun}) with respect to parameter $\bm{\uptheta}$, considering Eqn(\ref{eqn_gDL}), Eqn(\ref{eqn_OAL_BP}), and Eqn(\ref{eqn_OAL_JCBT}) yields
\begin{equation}
  \textbf{e}_y =  \mathbb{J}_\uptheta^T \delta\bm{\uptheta} = \Sigma_i (\mathbb{J}_{\uptheta}^{[i]})^T \delta\bm{\uptheta}^{[i]}=  \Sigma_i \Sigma_k (\mathbb{J}_{\uptheta_k}^{[i]})^T \delta\bm{\uptheta}_k^{[i]}
\label{eqn_ey_Taylor}
\end{equation}
or simply as
\[
  \textbf{e}_y = \mathbb{J}_{\uptheta_k^{[i]}}^T \delta\bm{\uptheta}_k^{[i]}
\]
according to Einstein notation for tensor calculation. In which
\begin{equation}
  \delta\bm{\uptheta} = \bm{\uptheta}^{*}-\bm{\uptheta}
  \label{eqn_DTheta}
\end{equation}
represents the deviation\footnote{Note that the conventional Taylor expansion uses $\delta\bm{\uptheta} = \bm{\uptheta}-\bm{\uptheta}^{*}$.} of the current parameter from the ideal targets $\bm{\uptheta}^{*}$ which captures the external world perfectly. Equivalent definitions apply to $\delta\bm{\uptheta}^{[i]}$ and $\delta\bm{\uptheta}_k^{[i]}$. Because the parameter adaptation process is driven directly by the error feedback flow of $\textbf{e}_y$ according to Eqn(\ref{eqn_OAL_GL_Update}), no further parameter update is generated in principle if $\textbf{e}_y = 0$. This equilibrium state follows from the minimum-energy law in Eqn(\ref{eqn_OAL_GAStb}). Under these conditions, Eqn(\ref{eqn_ey_Taylor}) reduces to\footnote{Note that $\mathbb{J}_{\uptheta}^{[i]}$ and $\textbf{J}_{\uptheta}^{[i]}$ refer to the global and local Jacobian sensitivity functions, respectively.}
\begin{equation}
    {\mathbb{J}_{\uptheta}^{[1]}}^T \delta\bm{\uptheta}^{[1]} + {\mathbb{J}_{\uptheta}^{[2]}}^T \delta\bm{\uptheta}^{[2]} + \cdots + {\mathbb{J}_{\uptheta}^{[i]}}^T \delta\bm{\uptheta}^{[i]}+ \cdots + {\mathbb{J}_{\uptheta}^{[n]}}^T \delta\bm{\uptheta}^{[n]} = \textbf{0}
\label{eqn_ey_TaylorSS}
\end{equation}
This demonstrates that the error feedback update law alone is insufficient to ensure global convergence toward $\delta\bm{\uptheta} = \textbf{0}$, both in terms of solution uniqueness and stability. Instead, it defines a high-dimensional manifold\footnote{The manifold dimension is high because $\delta\bm{\uptheta}^{[i]}$ is a matrix with size $n_i \times n_{i-1}$.} on which the resulting parameters may undergo stochastic sliding. Crucially, this equation indicates that the global parameter learning processes are strongly coupled through linear dependence.

\begin{figure}[h]
	\centering
	\includegraphics[scale=.40]{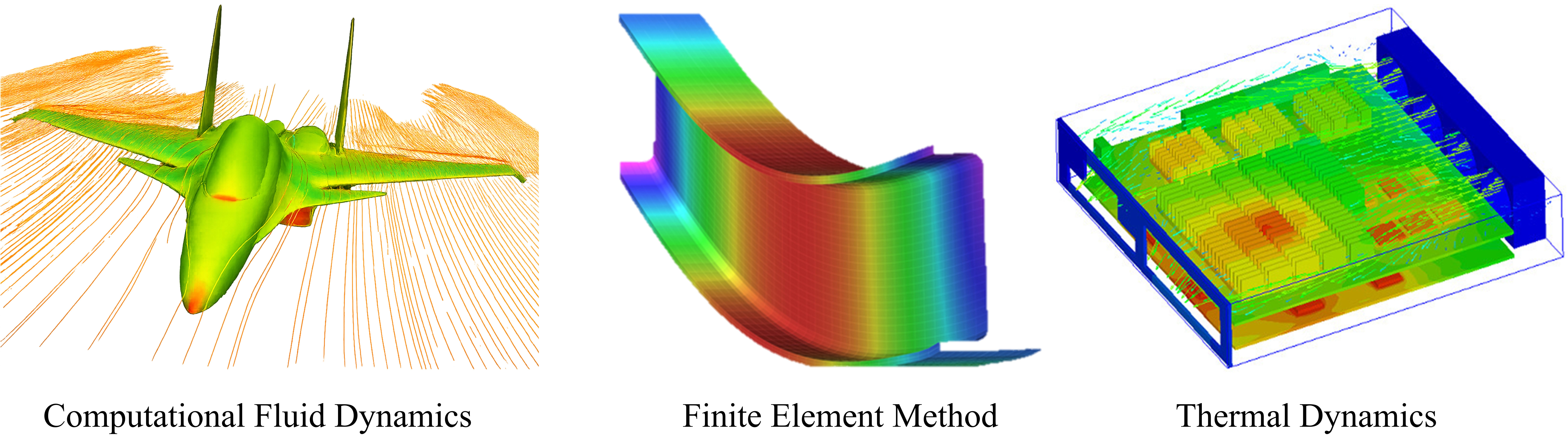}
	\caption{Structural Isomorphism of Large-Scale System Dynamics in Engineering}
	\label{fig:IsomorphismInNature}
\end{figure}

However, the structural \emph{isomorphism} of the feedback learning mechanism as given by Eqn(\ref{eqn_DL_BP}) implies that the same layer-wise argument can be applied across the parallel processing pathways. This observation does not by itself collapse the stationary manifold in Eqn(\ref{eqn_ey_TaylorSS}). Rather, it provides the structural template used below to analyze why additional excitation or damping is required to remove the residual zero dynamics.

The stability analysis is developed through an iterative deduction based on the structural isomorphism of the layered learning dynamics. The stability of such massive dynamical systems is also supported by empirical and physical analogies. In engineering, this is exemplified (see Fig.\ref{fig:IsomorphismInNature}) by Computational Fluid Dynamics (CFD), the Finite Element Method (FEM), and thermodynamics. These disciplines successfully capture huge-scale system dynamics precisely because the constituent elements share an identical structural framework.

This principle of underlying structural isomorphism mirrors fundamental processes in the natural world. For instance, macroscopic fluid mechanics and thermodynamics emerge from repeated local interactions among large numbers of individual molecules. Similarly, the macroscopic behavior governed by solid mechanics arises from collective interactions across vast crystalline lattices. These engineering paradigms and physical phenomena motivate the use of repeated local structure as an analytical simplification for ultra-large-scale dynamics.

\subsubsection*{Isomorphism of layered feedback learning}

\begin{figure}[h]
	\centering
	\includegraphics[scale=.67]{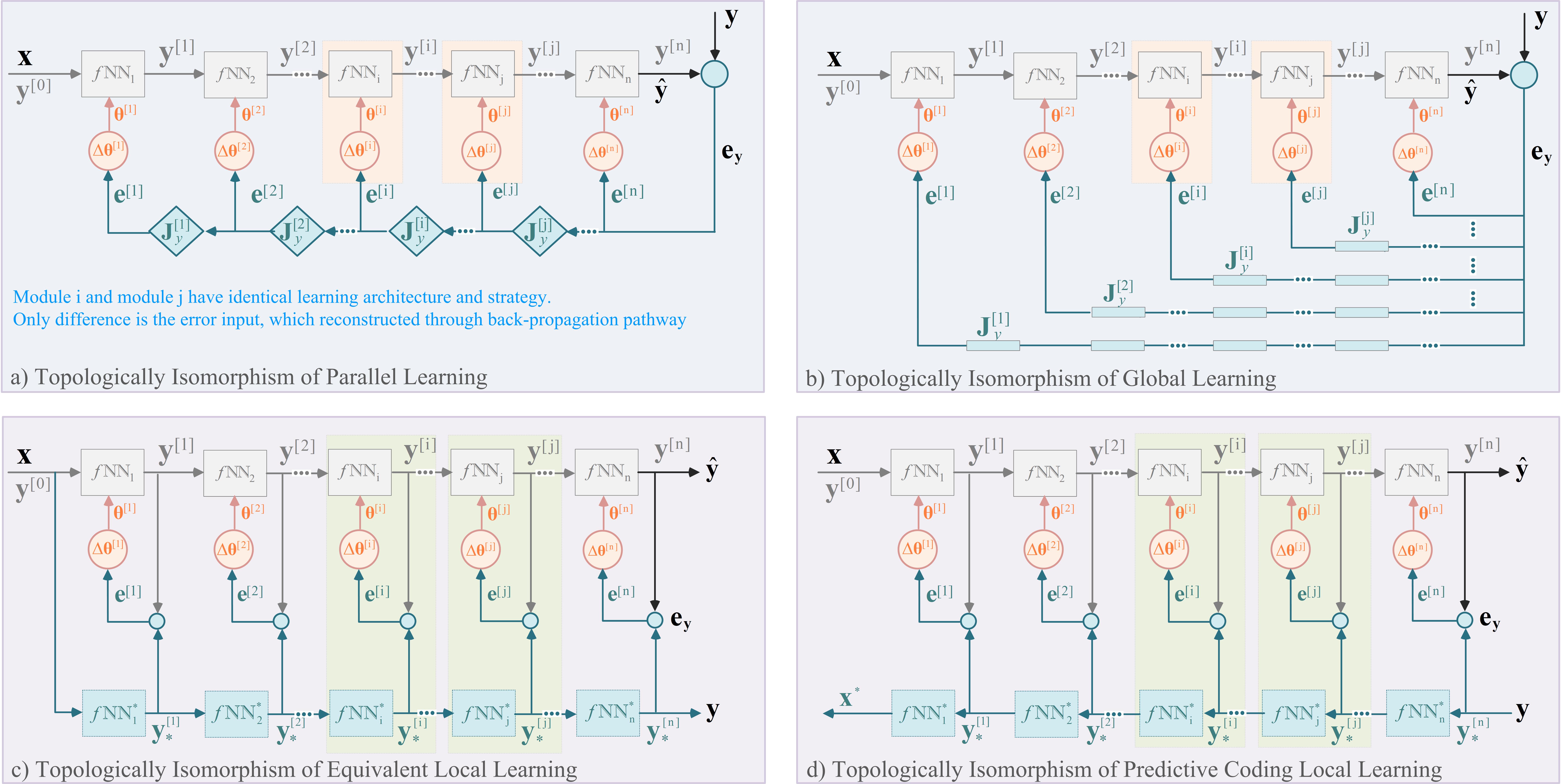}
	\caption{Topological Isomorphism of Feedback Learning}
	\label{fig:IsomorphismOAL}
\end{figure}

The feedback learning framework exhibits an identical learning topology across layers despite variations in latent-space dimensionality. Figure~\ref{fig:IsomorphismOAL} visualizes this topological isomorphism and supports the layer-wise analysis below.

\begin{itemize}
  \item \textbf{Layered Parallel Learning.} Each individual layer undergoes adaptive optimization driven by its respective layer-wise error $\mathbf{e}^{[i]}$, according to Eqn(\ref{eqn_OAL_Update}). For the stability analysis, the governing learning law is reformulated here in dynamic form.
      \begin{equation}
        \dot{\bm{\uptheta}}^{[i]}= \lambda_i \textbf{J}_\uptheta^{[i]} \mathbf{e}^{[i]}
        \label{eqn_Theta_Dyn}
      \end{equation}
      This law remains topologically and structurally invariant across all layers. Notably, the underlying manifold possesses high dimensionality, given that $\uptheta^{[i]}$ scales as an $n_i \times n_{i-1}$ matrix.
  \item \textbf{Learning Strategy Symmetry.} The optimization protocol displays strict layer-wise symmetry.
   The learning strategies for layer $i$, and $j$ are as, respectively
      \[
        \dot{\bm{\uptheta}}^{[i]}= \lambda_i \textbf{J}_\uptheta^{[i]} \mathbf{e}^{[i]}; \quad
        \dot{\bm{\uptheta}}^{[j]}= \lambda_j \textbf{J}_\uptheta^{[j]} \mathbf{e}^{[j]}
      \]
      Mathematically, these formulations are invariant under the permutation of indices \(i\) and \(j\), imbuing the ensemble of learning laws with group-theoretic properties. While exploiting the algebraic power of group analysis remains outside the scope of this study, it represents a promising avenue for future research;
  \item \textbf{Error Reconstruction Pathway.} The error propagation for any given layer is governed by
      \begin{equation}
        \mathbf{e}^{[i]} = \mathbb{J}_y^{[i]} \mathbf{e}_{y} = \prod_{k=i}^{n} \textbf{J}_y^{[k]} \mathbf{e}_{y}
        \label{eqn_ErrConstruct}
      \end{equation}
      The layered learning dynamics remain structurally coupled through error reconstruction. In particular, the dynamics of an early layer depend sequentially on the downstream layers.
\end{itemize}

To exploit the isomorphism topology more effectively, the parallel learning nature of the layered system is mapped into the architecture shown in Fig.\ref{fig:IsomorphismOAL}b, which is completely equivalent to Fig.\ref{fig:IsomorphismOAL}a. The two distinct features of this setup emerge immediately.
\begin{itemize}
  \item \textbf{Global Learning.} Every constituent layer executes its parameter learning driven directly by the terminal prediction error $\mathbf{e}_{y}$. Because of this long chain of information transformations, the error signals reaching early layers suffer from severe informational degradation and delay. Consequently, optimization for these layers is highly inefficient. Another critical consequence of this vulnerability is that newly integrated intermediate layers cannot adapt dynamically, as they are starved of localized, high-fidelity error gradients;
  \item \textbf{Cascade Learning.} The global routing topology gives rise to a large-scale multiloop optimization hierarchy. Under this configuration, downstream (late-stage) layers function as autonomous inner loops, while upstream (early-stage) layers operate as outer loops. This structural coupling is asymmetric. The inner loops continuously perturb the error spaces of the outer loops, while learning in the outer loops has almost no reciprocal effect on the inner loops.
\end{itemize}

Figures~\ref{fig:IsomorphismOAL}c and~\ref{fig:IsomorphismOAL}d present two local realizations of the same layered topology. In both cases, each layer is corrected through a locally reconstructed error signal rather than relying exclusively on the terminal prediction error.

\subsubsection*{Large-scale Multiple Loop Dynamics}
\textbf{Recursive Error State Dynamics.} Based on the global and cascaded learning structure, let the forward mapping of each layer be defined sequentially as:
\[
\mathbf{y}^{[i]} = f{\mathbb{NN}}_i (\mathbf{y}^{[i-1]}, \bm{\uptheta}^{[i]})
\]
which is essentially the combined version of Eqn(\ref{eqn_DLArch}). The parameter update law is governed by Eqn(\ref{eqn_Theta_Dyn}) as:
\[
 \dot{\bm{\uptheta}}^{[i]}= \lambda_i \textbf{J}_\uptheta^{[i]} \mathbf{e}^{[i]}
\]
where $\bm{\uptheta}^{[i]} \in \mathbb{R}^{n_i \times n_{i-1}}$. For a layer $i < n$, the localized error vector $\mathbf{e}^{[i]} \in \mathbb{R}^{n_i}$ is generated recursively by projecting the intermittent downstream error through the adjacent layer-to-layer state Jacobian matrix $\mathbf{J}_y^{[i]}$. This error vector $\mathbf{e}^{[i]}$ is given in Eqn(\ref{eqn_ErrConstruct}) as:
\[
 \mathbf{e}^{[i]} = \mathbb{J}_y^{[i]} \mathbf{e}_{y}
\]
The global tracking mismatch vector $\mathbf{e}_{y}$ driving the parameter learning is defined by Eqn(\ref{eqn_APErrFun}) as:
\[
 \mathbf{e}_y=\mathbf{y}_s-\hat{\mathbf{y}}_s
\]

\noindent \textbf{Cascaded Multi-Loop Dynamics.} To evaluate parameter convergence, the parameter error defined in Eqn({\ref{eqn_DTheta}}) is adopted as
\[
  \delta\bm{\uptheta} = \bm{\uptheta}^{*}-\bm{\uptheta}
\]
where $\bm{\uptheta}^{*}$ denotes the ideal parameter at equilibrium ($\dot{\bm{\uptheta}}^{*}=\mathbf{0}$), which changes only slowly by nature. Taking the time derivative, yields
\[
\delta\dot{\bm{\uptheta}} = -\dot{\bm{\uptheta}}
\]
Hence, the dynamic equation for parameter deviation at layer $i$ as
\begin{equation}
\delta\dot{\bm{\uptheta}}^{[i]} = - \lambda_i \mathbb{J}_\uptheta^{[i]} \mathbf{e}_y
\label{eqn_ParaErrDyn}
\end{equation}
Expanding this expression layer-by-layer, results in that:
\[\noindent\hspace{-5.5cm}
  \begin{array}{llll}

    \text{Layer n}   & \text{:}\qquad &
    \delta\dot{\bm{\uptheta}}^{[n]} = - \lambda_n \mathbb{J}_\uptheta^{[n]} \mathbf{e}_y
    &\quad=-\lambda_n \mathbf{J}_\uptheta^{[n]} \mathbf{e}_y\\
    \text{Layer j}   & \text{:}\qquad &
    \delta\dot{\bm{\uptheta}}^{[j]} = - \lambda_j \mathbb{J}_\uptheta^{[j]} \mathbf{e}_y
    &\quad= - \lambda_j \mathbf{J}_\uptheta^{[j]} \prod_{k=j}^{n} \textbf{J}_y^{[k]} \mathbf{e}_y\\
    \text{Layer i}   & \text{:}\qquad &
    \delta\dot{\bm{\uptheta}}^{[i]} = - \lambda_i \mathbb{J}_\uptheta^{[i]} \mathbf{e}_y
    &\quad= - \lambda_i \mathbf{J}_\uptheta^{[i]} \prod_{k=i}^{n} \textbf{J}_y^{[k]} \mathbf{e}_y\\
    \text{Layer 2}   & \text{:} \qquad&
    \delta\dot{\bm{\uptheta}}^{[2]} = - \lambda_{2} \mathbb{J}_\uptheta^{[2]} \mathbf{e}_y
    &\quad= - \lambda_{2} \mathbf{J}_\uptheta^{[2]} \prod_{k=2}^{n} \textbf{J}_y^{[k]} \mathbf{e}_y\\
    \text{Layer 1}   & \text{:} \qquad&
    \delta\dot{\bm{\uptheta}}^{[1]} = - \lambda_{1} \mathbb{J}_\uptheta^{[1]} \mathbf{e}_y
    &\quad= - \lambda_{1} \mathbf{J}_\uptheta^{[1]} \prod_{k=1}^{n} \textbf{J}_y^{[k]} \mathbf{e}_y
  \end{array}
\]

According to algebraic Eqn(\ref{eqn_ey_Taylor}), and introducing a diagonal error tracking factor $\mathbb{K}_y$, the dynamic version then becomes
\begin{equation}
   \dot{\textbf{e}}_y =
     {\mathbb{J}_{\uptheta}^{[1]}}^T \delta\dot{\bm{\uptheta}}^{[1]} +
     {\mathbb{J}_{\uptheta}^{[2]}}^T \delta\dot{\bm{\uptheta}}^{[2]} + \cdots + {\mathbb{J}_{\uptheta}^{[i]}}^T \delta\dot{\bm{\uptheta}}^{[i]} + \cdots + {\mathbb{J}_{\uptheta}^{[j]}}^T \delta\dot{\bm{\uptheta}}^{[j]} + \cdots + {\mathbb{J}_{\uptheta}^{[n]}}^T \delta\dot{\bm{\uptheta}}^{[n]} - \mathbb{K}_y \mathbf{e}_y
\label{eqn_dEy_Taylor}
\end{equation}

Stack the above vector differential equations, from layer $1$ and layer $2$, through layer $i$ and layer $j$, up to layer $n$, and combine the error dynamics described by Eqn(\ref{eqn_dEy_Taylor}), into a clean matrix format of $\dot{\mathbf{X}} = \mathbf{A}\mathbf{X}$, yields
\begin{equation}
  \resizebox{\textwidth}{!}{
  $\left[\begin{array}{c}
    \delta\dot{\bm{\uptheta}}^{[1]} \\
    \delta\dot{\bm{\uptheta}}^{[2]}\\
    \vdots\\
    \delta\dot{\bm{\uptheta}}^{[i]} \\
    \vdots\\
    \delta\dot{\bm{\uptheta}}^{[j]}\\
    \vdots\\
    \delta\dot{\bm{\uptheta}}^{[n]} \\
    \dot{\textbf{e}}_y
  \end{array}\right]=
  \left[\begin{array}{ccccccccc}
    \textbf{0} & \textbf{0} &  \cdots & \textbf{0} & \cdots & \textbf{0} & \cdots & \textbf{0} &
    - \lambda_1 \mathbf{J}_\uptheta^{[1]} \prod_{k=1}^{n} \textbf{J}_y^{[k]}\\
    \textbf{0} & \textbf{0} &  \cdots & \textbf{0} & \cdots & \textbf{0} & \cdots & \textbf{0} &
    - \lambda_2 \mathbf{J}_\uptheta^{[2]} \prod_{k=2}^{n} \textbf{J}_y^{[k]}\\
        \vdots & \vdots &  \vdots & \vdots & \vdots & \vdots & \vdots & \vdots & \vdots\\
    \textbf{0} & \textbf{0} &  \cdots & \textbf{0} & \cdots & \textbf{0} & \cdots & \textbf{0} &
    - \lambda_i \mathbf{J}_\uptheta^{[i]} \prod_{k=i}^{n} \textbf{J}_y^{[k]}\\
    \vdots & \cdots &  \vdots & \vdots & \vdots & \vdots & \vdots & \vdots & \vdots\\
    \textbf{0} & \textbf{0} &  \cdots & \textbf{0} & \cdots & \textbf{0} & \cdots & \textbf{0} &
    - \lambda_j \mathbf{J}_\uptheta^{[j]} \prod_{k=j}^{n} \textbf{J}_y^{[k]}\\
    \vdots & \cdots &  \vdots & \vdots & \vdots & \vdots & \vdots & \vdots & \vdots\\
    \textbf{0} & \textbf{0} &  \cdots & \textbf{0} & \cdots & \textbf{0} & \cdots & \textbf{0} &
     - \lambda_n \mathbf{J}_\uptheta^{[n]}\\
     (\mathbf{J}_\uptheta^{[1]} \prod_{k=1}^{n} \textbf{J}_y^{[k]})^T & (\mathbf{J}_\uptheta^{[2]} \prod_{k=2}^{n} \textbf{J}_y^{[k]})^T & \cdots & (\mathbf{J}_\uptheta^{[i]} \prod_{k=i}^{n} \textbf{J}_y^{[k]})^T & \cdots & (\mathbf{J}_\uptheta^{[j]} \prod_{k=j}^{n} \textbf{J}_y^{[k]})^T & \cdots & (\mathbf{J}_{\uptheta}^{[n]})^T & - \mathbb{K}_y\\
  \end{array}\right]
  \left[\begin{array}{c}
    \delta{\bm{\uptheta}}^{[1]} \vphantom{\delta\dot{\bm{\uptheta}}^{[1]}}\\
    \delta{\bm{\uptheta}}^{[2]} \vphantom{\delta\dot{\bm{\uptheta}}^{[2]}}\\
    \vdots\\
    \delta{\bm{\uptheta}}^{[i]} \vphantom{\delta\dot{\bm{\uptheta}}^{[i]}}\\
    \vdots\\
    \delta{\bm{\uptheta}}^{[j]} \vphantom{\delta\dot{\bm{\uptheta}}^{[j]}}\\
    \vdots\\
    \delta{\bm{\uptheta}}^{[n]} \vphantom{\delta\dot{\bm{\uptheta}}^{[n]}}\\
    {\textbf{e}}_y  \vphantom{\dot{\textbf{e}}_y}
  \end{array}\right]$
  }
 \label{eqn_sysDyn}
\end{equation}
Note that $\mathbf{A}$ is a block matrix, since the parameter state $\delta{\bm{\uptheta}}^{[i]}$ has size $n_i \times n_{i-1}$ and the error state $\textbf{e}_y$ has size $n_n \times 1$. A compact version of $\mathbf{A}$ is
\[
\mathbf{A} =
  \left[\begin{array}{ccccccccc}
    \textbf{0} & \textbf{0} &  \cdots & \textbf{0} & \cdots & \textbf{0} & \cdots & \textbf{0} &
    - \lambda_1 \mathbb{J}_\uptheta^{[1]} \\
    \textbf{0} & \textbf{0} &  \cdots & \textbf{0} & \cdots & \textbf{0} & \cdots & \textbf{0} &
    - \lambda_2 \mathbb{J}_\uptheta^{[2]} \\
    \vdots & \vdots &  \vdots & \vdots & \vdots & \vdots & \vdots & \vdots & \vdots\\
    \textbf{0} & \textbf{0} &  \cdots & \textbf{0} & \cdots & \textbf{0} & \cdots & \textbf{0} &
    - \lambda_i \mathbb{J}_\uptheta^{[i]} \\
    \vdots & \cdots &  \vdots & \vdots & \vdots & \vdots & \vdots & \vdots & \vdots\\
    \textbf{0} & \textbf{0} &  \cdots & \textbf{0} & \cdots & \textbf{0} & \cdots & \textbf{0} &
     - \lambda_j \mathbb{J}_\uptheta^{[j]} \\
    \vdots & \cdots &  \vdots & \vdots & \vdots & \vdots & \vdots & \vdots & \vdots\\
     \textbf{0} & \textbf{0} &  \cdots & \textbf{0} & \cdots & \textbf{0} & \cdots & \textbf{0} &
     - \lambda_n \mathbb{J}_\uptheta^{[n]}\\
    {\mathbb{J}_\uptheta^{[1]}}^T & {\mathbb{J}_\uptheta^{[2]} }^T & \cdots &  {\mathbb{J}_\uptheta^{[i]} }^T & \cdots & {\mathbb{J}_\uptheta^{[j]}}^T & \cdots &  {\mathbb{J}_{\uptheta}^{[n]}}^T & - \mathbb{K}_y\\
  \end{array}\right]
\]

This matrix representation reveals several characteristics of the system, providing additional insight from a dynamic point of view.
\begin{itemize}
  \item \textbf{P1: Zero Direct Causal Coupling.} The entire upper-left sub-matrix consists exclusively of zero blocks, demonstrating the absence of direct causal coupling between the layered parameter spaces. This decoupling is formally established by the following cross-sensitivity functions.
      \[
      \frac{\partial \delta\dot{\bm{\uptheta}}^{[i]}}{\partial \delta \bm{\uptheta}^{[j]}}=\mathbf{0} \quad (\forall i\ne j),\qquad
      \frac{\partial \delta \dot{\bm{\uptheta}}^{[i]}}{\partial \mathbf{e}_{y}}=-\lambda _{i}\mathbb{J}_{\uptheta }^{[i]}
      \]
      Clearly, the derivative of $\delta\bm{\uptheta}^{[i]}$ does not depend on the parameter state of $\delta\bm{\uptheta}^{[j]}$ of any other layer. This structural independence guarantees zero direct causal coupling between distinct layers. However, the layers remain implicitly interconnected through the tracking error feedback network. The parameter deviations $\delta\bm{\uptheta}$ of individual layers drive the reduction of the tracking error $\mathbf{e}_y$ collectively. Simultaneously, the emergent error $\mathbf{e}_{y}$ propagates back to govern the evolution of $\delta\bm{\uptheta}$ across all layers. A perturbation in $\delta\bm{\uptheta}^{[i]}$ can therefore influence $\delta\bm{\uptheta}^{[j]}$ only through the shared error state $\mathbf{e}_{y}$. This architecture effectively insulates the system against cross-destabilization within the internal parameter subspaces during adaptation.
  \item \textbf{P2: Skew-Symmetric Cross-Coupling.} The cross-coupling blocks of $\mathbf{A}$ are skew-symmetric up to the learning-rate scaling. Let \[
      \mathds{J}_{sk} = [\mathbb{J}_\uptheta^{[1]}, \mathbb{J}_\uptheta^{[2]}, \cdots, \mathbb{J}_\uptheta^{[i]}, \cdots, \mathbb{J}_\uptheta^{[n]}]^T
      \]
      and
      \[
      \Lambda = \text{diag}(\lambda_1 I, \lambda_2 I, \dots, \lambda_n I)
      \]
      Then $\mathbf{A}$ can be written in a more compact block form.
      \begin{equation}
      \mathbf{A}=\left[\begin{matrix}
      \mathbf{0}            & -\Lambda \mathds{J}_{sk}\\
      \mathds{J}_{sk}^{T}   & -\mathbb{K}_{y}
      \end{matrix}\right]
      \label{eqn_ACmpt}
      \end{equation}
      Scaled by the learning rates $\bm\lambda$, the cross-terms satisfy
      \(
      A_{12} = -\Lambda A_{21}^T
      \).
      This skew-symmetric type of matrix represents a passive system in general. Let a system given by $\dot{\bm{\Theta}} = \mathbb{A}_{sk}\bm{\Theta}$, with
      \[
        \mathbb{A}_{sk}=
        \left[\begin{matrix}
        \mathbf{0}            & -\Lambda \mathds{J}_{sk}\\
        \mathds{J}_{sk}^{T}   & \mathbf{0}
        \end{matrix}\right] \text{,} \qquad
        \mathbf{W} =
        \left[\begin{array}{cc}
         \bm\Lambda^{-1} & \mathbf{0} \\
         \mathbf{0}      & \mathbf{I} \\
         \end{array}\right] \text{,} \qquad
       \check{\mathbb{A}}_{sk} = \mathbf{W}\mathbb{A}_{sk} =
       \left[\begin{array}{cc}
           \mathbf{0}          & -\mathds{J}_{sk} \\
           \mathds{J}_{sk}^{T} & \mathbf{0} \\
       \end{array}\right]
      \]
      and introduce an energy function $\mathcal{L}_\Theta$ as
      \(
        \mathcal{L}_\Theta = \frac{1}{2}\bm{\Theta}^T\mathbf{W}\bm{\Theta}
      \),
      Then
      \begin{equation}
      \begin{array}{rccll}
       \dot{\mathcal{L}}_\Theta
       &=&   & {\bm\Theta}^T\mathbf{W}\dot{\bm{\Theta}} & \qquad \text{(\textbf{W} is positive definite)}\\
       &=&   & {\bm\Theta}^T\mathbf{W}\mathbb{A}_{sk}\bm{\Theta}&\\
       &=&   & {\bm\Theta}^T\check{\mathbb{A}}_{sk}\bm{\Theta} &\quad \text{Take a transpose}\\
       &=& - & {\bm\Theta}^T\check{\mathbb{A}}_{sk}\bm{\Theta} &\quad \text{Strictly skew-symmetric}\\
       &=&   & 0 & \quad x=-x \mapsto x=0
      \end{array}
      \label{eqn_Learn_Conservation}
      \end{equation}
      This implies that $\mathcal{L}_\Theta$ is constant. Thus, a skew-symmetric system is energy conservative and marginally stable. This is the case for parameter learning, and the associated dynamics of each layer are marginally stable.
  \item \textbf{P3: Strictly Damped Tracking Error Mode.} The last block $\mathbb{K}_{y}$ at bottom-right ensures that the global tracking error $\mathbf{e}_{y}$ experiences strict negative feedback, thereby facilitating system energy dissipation. With the energy function been defined as Eqn(\ref{eqn_Energy}), substituting Eqn(\ref{eqn_ParaErrDyn}) and Eqn(\ref{eqn_dEy_Taylor}), its derivative then becomes
      \begin{equation}
       \begin{array}{rcl}
        \dot{\mathcal{L}}_e &=& \textbf{e}_y^T \dot{\textbf{e}}_y\\
              &=& \textbf{e}_y^T
              \left({\mathbb{J}_{\uptheta}^{[1]}}^T \delta\dot{\bm{\uptheta}}^{[1]}
              + {\mathbb{J}_{\uptheta}^{[2]}}^T \delta\dot{\bm{\uptheta}}^{[2]} + \cdots
              + {\mathbb{J}_{\uptheta}^{[i]}}^T \delta\dot{\bm{\uptheta}}^{[i]} + \cdots
              + {\mathbb{J}_{\uptheta}^{[n]}}^T \delta\dot{\bm{\uptheta}}^{[n]}
              - \mathbb{K}_y \mathbf{e}_y\right)\\
              &=&  \textbf{e}_y^T
              \left(-\lambda_1{\mathbb{J}_{\uptheta}^{[1]}}^T \mathbb{J}_{\uptheta}^{[1]}
                     -\lambda_2{\mathbb{J}_{\uptheta}^{[2]}}^T \mathbb{J}_{\uptheta}^{[2]} - \cdots
                     -\lambda_i{\mathbb{J}_{\uptheta}^{[i]}}^T \mathbb{J}_{\uptheta}^{[i]} - \cdots
                     -\lambda_n{\mathbb{J}_{\uptheta}^{[n]}}^T \mathbb{J}_{\uptheta}^{[n]}
                     -\mathbb{K}_y\right) \textbf{e}_y\\
              &\leq& 0
       \end{array}
       \label{eqn_OAL_ErrStb}
      \end{equation}
      Introducing the damping gain $\mathbb{K}_y$ further enhances energy dissipation within the system. To illustrate this effect, the system matrix $\mathbf{A}$ is decomposed into a skew-symmetric component $\mathbb{A}_{sk}$ and a negative semi-definite damping block $\bm\Lambda_{_D} = \text{diag}(\mathbf{0}, -\mathbb{K}_y)$, where $\mathbb{K}_y \succ 0$. Define $\check{\mathbb{D}}_{sy}=\mathbf{W}\bm\Lambda_{_D}$. This decomposition yields
      \[
      \begin{array}{rccll}
       \dot{\mathcal{L}}_\Theta
       &=&   & {\bm\Theta}^T\mathbf{W}\dot{\bm{\Theta}}&\\
       &=&   & {\bm\Theta}^T\mathbf{W}\mathbb{A}_{sk}\bm{\Theta}
             + {\bm\Theta}^T\mathbf{W}\bm\Lambda_{_D}\bm{\Theta} &\\
       &=&   & \underbrace{{\bm\Theta}^T\check{\mathbb{A}}_{sk}\bm{\Theta}}
                          _{\text{Conservative } (= 0)}
               + \underbrace{{\bm\Theta}^T\check{\mathbb{D}}_{sy}\bm{\Theta}}
                          _{\text{Dissipative } (\le 0)}&\\
       &\leq&   & 0 &
      \end{array}
      \]
      Consequently, although the skew-symmetric term $\mathds{J}_{sk}$ reduces the energy associated with the tracking error, it merely redistributes into the parameter space to maintain a system level of energy balance. Ultimately, it is the injection of the damping matrix $\mathbb{K}_y$ that actively dissipates net energy from the system.
\end{itemize}

Properties \textbf{P1} and \textbf{P2} demonstrate that each learning loop outlined in Fig.~\ref{fig:IsomorphismOAL}b operates globally through the terminal layer. Specifically, inter-layer learning coupling occurs exclusively via the global tracking error, yielding marginal loop stability dictated by energy conservation. Conversely, Property \textbf{P3} establishes that each individual loop achieves asymptotic stability when isolated from the cascaded network and integrated with error tracking. From a dynamical systems perspective, this Online Active Learning (\textbf{OAL}) architecture can be interpreted as having a single stable outer loop (conventionally chosen as Layer 1), while all remaining loops operate as inner, marginally stable loops. This interpretation reveals a fundamental structural isomorphism: first, any arbitrary loop can topologically serve as the stable outer loop; second, all internal loops are strictly isomorphic with respect to both their learning topology and their marginal stability characteristics. Structurally, the outer loop dynamics govern the system's forced response, driving the tracking error asymptotically to zero under the control of the damping factor $\mathbb{K}_y$. In contrast, the internal loops define the unforced \emph{zero-dynamics} \cite{Isidori2013ZeroDyn} of the system, dictating the autonomous parameter trajectories that drift and freeze along a continuous equilibrium manifold once the tracking error vanishes.

\noindent \textbf{Spectral Analysis and Parameter Subspace Properties.} To evaluate the underlying learning dynamics, the eigenspace of matrix $\mathbf{A}$ is analyzed adopting the compact-block representation as Eqn(\ref{eqn_ACmpt}). Solving the characteristic equation $\mathbf{A}\mathbf{V} = \mu \mathbf{V}$ for the partitioned eigenvector $\mathbf{V} = [\mathbf{V}_\uptheta^T, \mathbf{V}_y^T]^T$ yields two coupled algebraic constraints as
\begin{align}
    -\bm\Lambda \mathds{J}_{sk} \mathbf{V}_y &= \mu \mathbf{V}_\uptheta \label{eqn_AVegn1} \\
    \mathds{J}_{sk}^T \mathbf{V}_\uptheta - \mathbb{K}_y \mathbf{V}_y &= \mu \mathbf{V}_y \label{eqn_AVegn2}
\end{align}
Eqn(\ref{eqn_AVegn1}) and Eqn(\ref{eqn_AVegn2}) reveals that the spectrum maps directly to the two functional sub-spaces as: 1) the stabilized space of the outer loop dynamic; and 2) the null-space of the inner zero dynamics.
\begin{itemize}
  \item \textbf{Stabilized Outer Loop Dynamics.} For the non-zero spectrum ($\mu \neq 0$) governing the transient tracking error, substituting $\mathbf{V}_\uptheta = -\frac{1}{\mu} \bm\Lambda \mathds{J}_{sk} \mathbf{V}_y$ from Eqn(\ref{eqn_AVegn1}) directly into Eqn(\ref{eqn_AVegn2}) isolates the outer loop as
         \begin{equation}
            \left(\mathds{J}_{sk}^T \bm\Lambda \mathds{J}_{sk}\right) \mathbf{V}_y = -\mu (\mu \mathbf{I} + \mathbb{K}_y) \mathbf{V}_y
         \label{eqn_OuterCore}
         \end{equation}
      Let $\sigma^2 > 0$ denote an eigenvalue of the symmetric and positive-definite matrix $\mathbf{M}_c = \mathds{J}_{sk}^T \bm\Lambda \mathds{J}_{sk}$, which represents the aggregate, learning-rate-scaled sensitivity of the cascaded layers. Assuming a scalar configuration for the damping factor $\mathbb{K}_y = k_y \mathbf{I}$, this relationship collapses to a scalar quadratic characteristic equation:
      \[
         \mu^{2} + k_{y}\mu + \sigma^{2} = 0
      \]
      This proves that the forced response of the outer tracking loop is stabilized, since the real parts of these eigenvalues are strictly negative, $\text{Re}(\mu) < 0$, as a result of the designed damping factor $k_y > 0$. Furthermore, the system can be explicitly designed to exhibit over-damped ($k_y^2 \geq 4\sigma^2$) characteristics if required.
    \item \textbf{Zero Dynamics of the Inner Null-Space.} From a nonlinear dynamics perspective, evaluating the null-spectrum ($\mu = 0$) characterizes the unforced \emph{zero-dynamics} \cite{Isidori2013ZeroDyn} of the \textbf{OAL} system, specifically representing the internal parameter adaptations when the tracking error is held to zero. Setting $\mu = 0$ in Eqn(\ref{eqn_AVegn1}) directly dictates that the tracking error component vanishes ($\mathbf{V}_y = \mathbf{0}$). Substituting this into Eqn(\ref{eqn_AVegn2}) formalizes the zero-dynamics eigenspace as
         \begin{equation}
            \mathbf{V}_{0} = [\mathbf{V}_{\uptheta}, \mathbf{0}]^T \quad \text{subject to} \quad \mathds{J}_{sk}^T \mathbf{V}_\uptheta = \mathbf{0}
         \label{eqn_ZeroCore}
         \end{equation}
       Consequently, the parametric zero dynamics locally reside within the null space of $\mathds{J}_{sk}$. Because the collective parameter space is overwhelmingly large relative to the tracking output space, $\mathds{J}_{sk}$ fundamentally possesses a high-dimensional kernel, namely $\ker(\mathds{J}_{sk})$. Without an additional mechanism to break this structural symmetry, internal parameter updates stall along the manifold explicitly characterized by Eqn(\ref{eqn_ey_TaylorSS}). Thus, when the outer loop collapses to zero tracking error, the internal zero dynamics are only marginally stable. This permanently locks the learnable parameters into an infinite number of sub-optimal combinations within the just-mentioned null space.
\end{itemize}

\subsubsection*{Internal Zero Dynamics Stabilization}

To guarantee global parameter convergence where $\delta\bm{\uptheta}^{[i]}(t)\rightarrow \mathbf{0}$ as $t\rightarrow \infty$ across all layers $i$, the system must disrupt the stationary high-dimensional manifold defined by Eqn(\ref{eqn_ey_TaylorSS}) that resides within the kernel of $\mathbb{A}_{sk}$. This intervention fundamentally alters the system properties, forcing the internal zero-dynamics to shift toward asymptotic stability via the Persistency of Excitation (PE) condition.

\noindent \textbf{Persistency of Excitation Method.} Within the \textbf{OAL} framework, the dynamic streaming of input signals forces the composite cascaded Jacobian matrix $\mathds{J}_{sk}(t)$ to evolve as a time-dependent component. Specifically, if $\mathds{J}_{sk}(t)$ satisfies the classical PE constraint:
\begin{equation}
    \int_{t}^{t+T} \mathds{J}_{sk}^{T}(\tau ) \mathds{J}_{sk}^{}(\tau ) \, d\tau \ge \alpha \mathbf{I} > 0
\label{eqn_PE_Condition}
\end{equation}
for $T > 0$ and $\alpha > 0$. This condition effectively rotates the high-dimensional kernel across the parameter space, preventing the state trajectories from settling into sub-optimal equilibria. Thus, parametric adaptation remains active until the error vector is explicitly driven to $\delta\bm{\uptheta} = \mathbf{0}$. Importantly, this requirement is stronger than pointwise non-vanishing Jacobians. The local sensitivities must vary over the finite horizon so that their accumulated Gramian is full rank; otherwise, the learning trajectory may still freeze in a fixed null direction.

Dynamically, satisfying the PE condition removes the residual zero-dynamics over a finite time window. By strengthening the energy argument from pointwise non-increase, $\dot{\mathcal{L}} \le 0$, to finite-window contraction, this mechanism dissipates the residual energy associated with the null space. Instead of allowing parameters to drift along a flat, marginally stable valley, sufficient excitation prevents the output error from remaining zero unless the parameter errors also vanish. This upgrades the internal learning dynamics from marginal stability to asymptotic parameter convergence under the PE condition.

\noindent \textbf{Active Damping Approach.} Although the PE condition can remove the residual zero dynamics, it remains a passive requirement that depends entirely on external excitation. When input signals lack sufficient variation, parameter adaptation can still become trapped within the null space. To counteract these situations, active damping for parameter learning, as introduced in Eqn(\ref{eqn_OAL_LearnDyn_R}), Eqn(\ref{eqn_OAL_LearnDyn_L}), and Eqn(\ref{eqn_OAL_LearnDyn_P}), is injected to dissipate the energy associated with internal parametric fluctuations. This intervention introduces a dedicated leakage matrix, $\mathbb{K}_\uptheta^{[i]} > 0$, directly into the diagonal blocks of the state-space matrix $\mathbf{A}$ from Eqn(\ref{eqn_sysDyn}). Under this configuration, the parameters follow the augmented dynamic update law:
\begin{equation}
  \dot{\bm{\uptheta}}^{[i]}= - \mathbb{K}_\uptheta^{[i]} \bm{\uptheta}^{[i]} + \lambda_i \mathbb{J}_\uptheta^{[i]} \mathbf{e}_y
  \label{eqn_Theta_DynNew}
\end{equation}
which serves as the regularized counterpart to the nominal dynamics in Eqn(\ref{eqn_Theta_Dyn}). From an algorithmic perspective, the discrete-time implementation of this iterative structure closely mirrors weight decay in standard optimization. This mechanism alters the residual zero dynamics of the inner loops by introducing strict energy dissipation into the kernel of the composite Jacobian $\mathds{J}_{sk}$. Consequently, even when the tracking error vanishes ($\mathbf{e}_y = \mathbf{0}$), the autonomous parameter trajectories are pulled toward a bounded region, preventing parametric drift. From an offline-training perspective, this damping acts as weight decay or $L_2$-style regularization, rather than parameter normalization. Whereas momentum accelerates offline convergence \cite{qian1999momentum}, online damping stabilizes the adaptive loop against excitation deficits.

\subsubsection*{Cascade Convergence and Stability Analysis}

\textbf{State-Space Representation.} Incorporating the active damping matrices for parameter learning as diagonal elements across the sequential layer blocks yields the following updated version of Eqn(\ref{eqn_sysDyn}) as
\begin{equation}
\begin{bmatrix}
    \delta\dot{\bm{\uptheta}}^{[1]} \\
    \delta\dot{\bm{\uptheta}}^{[2]} \\
    \vdots \\
    \delta\dot{\bm{\uptheta}}^{[i]} \\
    \vdots \\
    \delta\dot{\bm{\uptheta}}^{[n]} \\
    \dot{\mathbf{e}}_{y}
\end{bmatrix} =
\begin{bmatrix}
    -\mathbb{K}_\uptheta^{[1]} & \mathbf{0} & \dots & \mathbf{0} & \dots & \mathbf{0} & -\lambda_{1}\mathbb{J}_{\uptheta}^{[1]} \\
    \mathbf{0} & -\mathbb{K}_\uptheta^{[2]} & \dots & \mathbf{0} & \dots & \mathbf{0} & -\lambda_{2}\mathbb{J}_{\uptheta}^{[2]} \\
    \vdots & \vdots & \ddots & \vdots & \ddots & \vdots & \vdots \\
    \mathbf{0} & \mathbf{0} & \dots & -\mathbb{K}_\uptheta^{[i]} & \dots & \mathbf{0} & -\lambda_{i}\mathbb{J}_{\uptheta}^{[i]} \\
    \vdots & \vdots & \ddots & \vdots & \ddots & \vdots & \vdots \\
    \mathbf{0} & \mathbf{0} & \dots & \mathbf{0} & \dots & -\mathbb{K}_\uptheta^{[n]} & -\lambda_{n}\mathbb{J}_{\uptheta}^{[n]} \\
    {\mathbb{J}_{\uptheta}^{[1]}}^{T} & {\mathbb{J}_{\uptheta}^{[2]}}^{T} & \dots & {\mathbb{J}_{\uptheta}^{[i]}}^{T} & \dots & {\mathbb{J}_{\uptheta}^{[n]}}^{T} & -\mathbb{K}_{y}
\end{bmatrix}
\begin{bmatrix}
    \delta \bm{\uptheta}^{[1]} \vphantom{\delta\dot{\bm{\uptheta}}^{[1]}}\\
    \delta \bm{\uptheta}^{[2]} \vphantom{\delta\dot{\bm{\uptheta}}^{[2]}}\\
    \vdots \\
    \delta \bm{\uptheta}^{[i]} \vphantom{\delta\dot{\bm{\uptheta}}^{[i]}}\\
    \vdots \\
    \delta \bm{\uptheta}^{[n]} \vphantom{\delta\dot{\bm{\uptheta}}^{[n]}}\\
    \mathbf{e}_{y} \vphantom{\dot{\mathbf{e}}_{y}}
\end{bmatrix}
\label{eqn_sysDyn_new}
\end{equation}
where $\mathbb{K}_\uptheta^{[i]} > 0$ denotes the active damping for Layer $i$, and $\mathbb{J}_{\uptheta}^{[i]}$ denotes the lumped Jacobian matching the cascaded structural topology. To establish global asymptotic stability across this multi-loop architecture, the topological isomorphism method is adopted. Because the parameter learning topology is identical across all layers, evaluating the stability of a generalized Layer $i$ from a dynamical systems perspective functionally validates the stability properties of the entire unified system.

\noindent \textbf{Parameter Learning Local Stability.}  To establish the learning stability, layer $i$ is isolated locally. All upstream and downstream parameter errors ($\delta\bm{\uptheta}^{[1]}, \dots, \delta\bm{\uptheta}^{[k]}, \dots, \delta\bm{\uptheta}^{[n]}$, for $k \neq i$) are treated as bounded external disturbances. A local Lyapunov-like energy function $\mathcal{L}_i(t)$ is defined as
\begin{equation}
    \mathcal{L}_i(t) = \frac{1}{2}\mathbf{e}_{y}^{T}\mathbf{e}_{y} + \frac{1}{2\lambda_{i}}{\delta \bm{\uptheta}^{[i]}}^{T}\delta \bm{\uptheta}^{[i]}
    \label{eqn_Li_Energy}
\end{equation}
The derivative of $\mathcal{L}_i(t)$ along the state trajectories is
\[
    \dot{\mathcal{L}}_{i}(t) = \mathbf{e}_{y}^{T}\dot{\mathbf{e}}_{y} + \frac{1}{\lambda_{i}}{\delta \bm{\uptheta}^{[i]}}^{T}\delta \dot{\bm{\uptheta}}^{[i]}
\]
Substituting the corresponding row dynamics from \eqref{eqn_sysDyn_new} for $\dot{\mathbf{e}}_{y}$ and $\delta\dot{\bm{\uptheta}}^{[i]}$ yields:
\[
\begin{aligned}
    \dot{\mathcal{L}}_{i}(t) &= -\mathbf{e}_{y}^{T}\mathbb{K}_{y}\mathbf{e}_{y} + \mathbf{e}_{y}^{T}
    \left( \sum\limits_{\substack{k=1,k\neq i}}^{n} {\mathbb{J}_{\uptheta}^{[k]}}^{T}\delta\bm{\uptheta}^{[k]} + {\mathbb{J}_{\uptheta}^{[i]}}^{T}\delta \bm{\uptheta}^{[i]} \right)
    - \frac{1}{\lambda_{i}} {\delta \bm{\uptheta}^{[i]}}^{T} \left(\lambda_{i}\mathbb{J}_{\uptheta}^{[i]}\mathbf{e}_{y} + \mathbb{K}_\uptheta^{[i]} \delta \bm{\uptheta}^{[i]}\right) \\
    &= -\mathbf{e}_{y}^{T}\mathbb{K}_{y}\mathbf{e}_{y} - \frac{1}{\lambda_{i}} {\delta \bm{\uptheta}^{[i]}}^{T}\mathbb{K}_\uptheta^{[i]}\delta\bm{\uptheta}^{[i]} + \mathbf{\check{\Delta}}_{i}(t)
\end{aligned}
\]
where the cross-layer coupling term $\mathbf{\check{\Delta}}_{i}(t)$ is defined as:
\[
    \mathbf{\check{\Delta}}_{i}(t) = \mathbf{e}_{y}^{T}(t) \sum\limits_{\substack{k=1,k\neq i}}^{n} {\mathbb{J}_{\uptheta}^{[k]}}^{T}\delta\bm{\uptheta}^{[k]}
\]
This term represents the multi-layer disturbance generated by the residual learning errors of the remaining cascaded layers. For an isolated layer, $\mathbf{\check{\Delta}}_{i}(t)$ must therefore be treated as an external coupling term rather than eliminated locally. The local negative-definite part is
\begin{equation}
    \dot{\mathcal{L}}_{i}(t) - \mathbf{\check{\Delta}}_{i}(t) = -\left( \mathbf{e}_{y}^{T}\mathbb{K}_{y}\mathbf{e}_{y} + \frac{1}{\lambda_{i}} {\delta \bm{\uptheta}^{[i]}}^{T}\mathbb{K}_\uptheta^{[i]} \delta \bm{\uptheta}^{[i]} \right) \leq 0
 \label{eqn_lcLyp_SBT}
\end{equation}
If the residual coupling is absent, or after the complete cascaded cancellation shown in the next subsection is applied, the derivative is non-positive and vanishes only when both the tracking error and the local parameter error vanish ($\mathbf{e}_{y} = \mathbf{0}$ and $\delta\bm{\uptheta}^{[i]} = \mathbf{0}$). Thus, this local calculation identifies the damped layer-wise stability mechanism, while the full asymptotic conclusion requires the global cascaded proof below.

From a physical perspective, $\mathbb{K}_{y}$ acts as a virtual damping that dissipates energy within the tracking loop (which represents a topological rather than a physical loop), while $\mathbb{K}_\uptheta^{[i]}$ mitigates the energy associated with the parameter adaptation loop of layer $i$. Notably, the presence of the adaptation gain $\mathbb{K}_\uptheta^{[i]}$ renders the tracking gain $\mathbb{K}_{y}$ non-essential, thereby providing enhanced control design flexibility.

\noindent \textbf{Parameter Learning Cascaded Stability.} Because the parameter learning of all layers is coupled through the global error tracking loop, a comprehensive assessment of the integrated system is necessary. This evaluation must simultaneously account for tracking error stability, parameter convergence, and cross-layer learning interactions. To this end, a global Lyapunov candidate function $\mathcal{L}(t)$ is defined, which is intrinsically a multi-layer extension of \eqref{eqn_Energy},
\begin{equation}
    \mathcal{L}(\mathbf{y},\bm{\uptheta},t) = \frac{1}{2}\mathbf{e}_{y}^{T}\mathbf{e}_{y} + \sum_{i=1}^{n}\frac{1}{2\lambda_{i}} {\delta \bm{\uptheta}^{[i]}}^{T}\delta \bm{\uptheta}^{[i]}
    \label{eqn_Energy_OverAll}
\end{equation}
Differentiating $\mathcal{L}(t)$ along the state trajectories of the full cascaded space yields:
\[
    \dot{\mathcal{L}}(t) = \mathbf{e}_{y}^{T}\dot{\mathbf{e}}_{y} + \sum_{i=1}^{n}\frac{1}{\lambda_{i}}{\delta \bm{\uptheta}^{[i]}}^{T} \delta\dot{\bm{\uptheta}}^{[i]}
\]
Substituting the full system dynamics from \eqref{eqn_sysDyn_new} expands the time derivative into
\[
 \begin{aligned}
    \dot{\mathcal{L}}(t) &= -\mathbf{e}_{y}^{T}\mathbb{K}_{y}\mathbf{e}_{y} + \mathbf{e}_{y}^{T} \sum_{i=1}^{n}{\mathbb{J}_{\uptheta}^{[i]}}^{T}\delta \bm{\uptheta}^{[i]} - \sum_{i=1}^{n}\frac{1}{\lambda_{i}} {\delta\bm{\uptheta}^{[i]}}^{T}\left(\lambda_{i}\mathbb{J}_{\uptheta}^{[i]}\mathbf{e}_{y} + \mathbb{K}_\uptheta^{[i]} \delta \bm{\uptheta}^{[i]}\right) \\
    &= -\mathbf{e}_{y}^{T} \mathbb{K}_{y} \mathbf{e}_{y} + \sum_{i=1}^{n} \mathbf{e}_{y}^{T}{\mathbb{J}_{\uptheta}^{[i]}}^{T} \delta \bm{\uptheta}^{[i]} - \sum_{i=1}^{n} {\delta\bm{\uptheta}^{[i]}}^{T}\mathbb{J}_{\uptheta}^{[i]} \mathbf{e}_{y} - \sum_{i=1}^{n}\frac{1}{\lambda_{i}} {\delta\bm{\uptheta}^{[i]}}^{T}\mathbb{K}_\uptheta^{[i]} \delta \bm{\uptheta}^{[i]}
 \end{aligned}
\]
Noting that the cross-layer interaction terms are scalars, they satisfy the transpose identity
\[
\left(\sum_{i=1}^{n} \mathbf{e}_{y}^{T}{\mathbb{J}_{\uptheta}^{[i]}}^{T} \delta \bm{\uptheta}^{[i]}\right)^T = \sum_{i=1}^{n} {\delta\bm{\uptheta}^{[i]}}^{T}\mathbb{J}_{\uptheta}^{[i]} \mathbf{e}_{y}
\]
Consequently, these skew-symmetric coupling terms cancel out identically. This cancellation implies that the learning adaptation law perfectly matches and neutralizes the dynamic interactions across the cascaded layers, reducing the derivative to
\begin{equation}
  \dot{\mathcal{L}}(t) = -\mathbf{e}_{y}^{T}\mathbb{K}_{y}\mathbf{e}_{y} - \sum_{i=1}^{n}\frac{1}{\lambda_{i}} {\delta\bm{\uptheta}^{[i]}}^{T}\mathbb{K}_\uptheta^{[i]} \delta \bm{\uptheta}^{[i]} \leq 0
  \label{eqn_GlLyp_SBT}
\end{equation}
Eqn(\ref{eqn_GlLyp_SBT}) establishes that $\mathcal{L}(t)$ is monotonically non-increasing and bounded from below ($\mathcal{L}(t) \ge 0$). By invoking Barbalat's lemma \cite{farkas2016variations} alongside Lyapunov stability criteria \cite{nguyen2018lyapunov,barkana2014invariance}, it follows that $\lim_{t \to \infty} \dot{\mathcal{L}}(t) = 0$, implying that the system trajectories eventually converge to a minimal energy state \cite{friston2010Brain}. Because the coefficient matrices $\mathbb{K}_{y}$ and $\mathbb{K}_\uptheta^{[i]}$, $i \in \{1,2, \dots, n\}$, of the two remaining quadratic forms in Eqn(\ref{eqn_GlLyp_SBT}) are strictly positive definite, it follows that
\[
    \lim_{t\rightarrow \infty }\mathbf{e}_{y}(t) = \mathbf{0} \quad \text{and} \quad \lim_{t\rightarrow \infty }\delta \bm{\uptheta}^{[i]}(t) = \mathbf{0} \quad \forall i \in \{1,2, \dots, n\}
\]
This completes the proof of asymptotic stability for the cascaded \textbf{OAL} learning dynamics. In particular, Eqn\eqref{eqn_GlLyp_SBT} again shows that the tracking gain \(\mathbb{K}_{y}\) is non-essential, justifying its frequent omission in practical implementations.

However, the active damping term $\mathbb{K}_\uptheta^{[i]}$ also governs how fast the parameter energy decays. To quantify this effect without loss of generality, consider a simplified scenario in which the tracking gain $\mathbb{K}_{y}$ is omitted and the damping matrix $\mathbb{K}_\uptheta^{[i]}$ reduces to a uniform scalar factor $\kappa_i = \kappa$. The energy function in Eqn(\ref{eqn_Energy_OverAll}) then reduces to
\[
    \mathcal{L}(t) = \sum_{i=1}^{n}\frac{1}{2\lambda_{i}} {\delta \bm{\uptheta}^{[i]}}^{T}\delta \bm{\uptheta}^{[i]}
\]
and its derivative, which is a simplified version of Eqn(\ref{eqn_GlLyp_SBT}), becomes
\[
 \begin{aligned}
  \dot{\mathcal{L}}(t)
  &= - \sum_{i=1}^{n}\frac{\kappa_i}{\lambda_{i}} {\delta\bm{\uptheta}^{[i]}}^{T} \delta \bm{\uptheta}^{[i]} \\
  &= - 2\kappa \sum_{i=1}^{n} \frac{1}{2\lambda_{i}} \|{\delta\bm{\uptheta}^{[i]}}\|{}_{2}^{2}\\
  &= - 2\kappa  \mathcal{L}(t)
 \end{aligned}
\]
The temporal profile of this simplified system energy is governed by the exact analytical solution
\[
\mathcal{L}(t) = \mathcal{L}(0)e^{-2\kappa t}
\]
This expression shows that the damping factor $\kappa$ sets the exponential decay rate of the parameter energy. Larger values produce faster decay in this simplified model, although practical tuning must balance retention of learned structure against adaptation to new data.

\subsubsection*{Overall OAL learning Stability}

While the preceding subsection establishes the asymptotic stability of the cascaded learning framework, overall system convergence still depends on two foundational conditions. First, the ideal parameter must be unique to guarantee well-defined convergence. In particular, this uniqueness refers not to a static parameter state $\bm{\uptheta}^*$, but to a time-varying trajectory $\bm{\uptheta}^*(t)$. The trajectory remains unique at each time $t \in \mathbb{R}^+$ while encoding environmental dynamics over a given interval. Second, within each layer $i \in \{1, 2, \dots, n\}$, the parameter matrix $\bm{\uptheta}^{[i]}$ regulates multiple concurrent inference channels. Because cascaded inter-layer coupling is not inherently isomorphic to intra-layer coupling, these parallel learning processes must independently satisfy localized stability criteria to ensure global convergence of the multilayer system.

\noindent \textbf{Parameter Learning Global Stability.} The state-space Eqn\eqref{eqn_sysDyn_new} characterizes the system dynamics exclusively via a localized linearization, meaning the associated lumped Jacobian matrix $\mathbb{J}_\uptheta^{[i]}$ captures only the local behaviors. To extend the cascaded learning convergence globally, these Jacobians cannot merely serve as local derivatives. Instead, the underlying global mapping
\begin{equation}
  \bm{\mathcal{F}}_\uptheta^{[i]}:\:\quad \mathfrak{M}_{\mathbf{y}^{[n]}} \to \mathfrak{M}_{\bm{\uptheta}^{[i]}}
 \label{eqn_y2tha_map}
\end{equation}
must constitute a global monotonic isomorphism. Specifically, because this mapping is a cascaded composition propagating sequentially through the intermediate latent spaces \(\mathfrak{M}_{\mathbf{y}^{[n-1]}} \times \cdots \times \mathfrak{M}_{\mathbf{y}^{[i]}}\), it must establish a smooth global diffeomorphism across the entire joint topological space. This condition requires that the lumped Jacobian $\mathbb{J}_\uptheta^{[i]}$ remains non-singular and strictly sign-preserving across the entire product space, defined as
\[
\bm{\Omega}^{[i]}:= \; \mathfrak{M}_{\mathbf{y}^{[n]}} \times \mathfrak{M}_{\mathbf{y}^{[n-1]}} \times \cdots \times \mathfrak{M}_{\mathbf{y}^{[i]}} \times \mathfrak{M}_{\bm{\uptheta}^{[i]}}
\]

The geometric deduction of the global diffeomorphism requirement to the uniqueness of the ideal trajectory $\bm{\uptheta}^*(t)$ is fundamentally rooted in the topological architecture of the system energy function. Referring to the primary formulation of Eqn.\eqref{eqn_Energy}, the energy function $\mathcal{L}_e\big(\bm{\uptheta}^{[i]}\big)$ to be minimized is explicitly formulated to isolate the dependency on the layer-wise parameter matrix $\bm{\uptheta}^{[i]}(t)$:
\[
 \begin{array}{rcl}
  \mathcal{L}_e(\bm{\uptheta}^{[i]}) &=& \frac{1}{2} \| \mathbf{y}(t) - f{\mathbb{NN}}_\mathrm{dl}^{[i]}[\bm{\uptheta}^{[i]}(t)]\|^2.\\
  &=&\frac{1}{2} \| \mathbf{y}(t) -  f{\mathbb{NN}}_n \circ \cdots \circ f{\mathbb{NN}}_{i-1} \circ f{\mathbb{NN}}_i [\bm{\uptheta}^{[i]}(t)]  \|^2
 \end{array}
\]
This defines the mapping $\mathcal{L}_e: \mathfrak{M}_{\bm{\uptheta}^{[i]}} \times \mathfrak{M}_{\mathbf{y}^{[n]}} \to \mathbb{R}^+$, representing the energy landscape under a squared Euclidean distance metric.

Because $\mathcal{L}_e^i$ prescribes a strictly convex parabolic geometry with respect to the prediction error $\mathbf{e}_y$, the underlying topology of the ideal parameter $\bm{\uptheta}^{[i]}_*$ is dictated entirely by the properties of the neural mapping $f{\mathbb{NN}}_\mathrm{dl}^{[i]}$. The first-order necessary condition for optimality requires the gradient to vanish identically at the target trajectory, yielding
\begin{equation}
   \nabla_{\bm{\uptheta}^{[i]}} \mathcal{L} \;\Big|_{\bm{\uptheta}^{[i]} = \bm{\uptheta}^{[i]}_*} = -{\mathbb{J}_{\uptheta}^{[i]}}^T
   \left(
   \mathbf{y} - f{\mathbb{NN}}_\mathrm{dl}^{[i]}[\bm{\uptheta}^{[i]}(t)]
   \right)\Big|_{\bm{\uptheta}^{[i]} = \bm{\uptheta}^{[i]}_*}
   = \mathbf{0}
\label{eqn_UqnSln}
\end{equation}
Consequently, guaranteeing the global uniqueness of the instantaneous target $\bm{\uptheta}^{[i]}_*(t)$ strictly necessitates two core conditions:
\begin{itemize}
    \item[\textbf{C1.}] The lumped Jacobian must be non-singular, $\mathbb{J}_{\uptheta}^{[i]} \neq \mathbf{0}$. This implicitly requires the sign-invariance of $\mathbb{J}_{\uptheta}^{[i]}$, as a change in sign within the operational domain would topologically lead to $\mathbb{J}_{\uptheta}^{[i]} = \mathbf{0}$ via the intermediate value theorem;
    \item[\textbf{C2.}] The deep mapping $f{\mathbb{NN}}_\mathrm{dl}^{[i]}$ must be strictly monotonic with respect to $\bm{\uptheta}^{[i]}$. This condition is vital to prevent geometric folding or multi-branched overlapping of the parameter-to-output manifold.
\end{itemize}

Given that the global network mapping
\begin{equation}
  f{\mathbb{NN}}_\mathrm{dl}^{[i]}:\: \quad  \mathfrak{M}_{\bm{\uptheta}^{[i]}}\to \mathfrak{M}_{\mathbf{y}^{[n]}}
  \label{eqn_tha2y_map}
\end{equation}
is a composition of functions $f{\mathbb{NN}}_i, f{\mathbb{NN}}_{i+1}, \cdots, f{\mathbb{NN}}_{n}$, condition \textbf{C2} is satisfied by ensuring each constitutive layered mapping exhibits strict global monotonicity. Under these constraints, the Euclidean energy surface is guaranteed to preserve a singular, globally attractive unique minimum across the entire domain.

Thus, $\bm{\mathcal{F}}_\uptheta^{[i]}$ as Eqn.\eqref{eqn_y2tha_map} and $f{\mathbb{NN}}_\mathrm{dl}^{[i]}$ as Eqn.\eqref{eqn_tha2y_map} are revealed to exhibit an intrinsic structural duality. According to Hadamard-Banach-Mazur theorem \cite{plastock1974homeomorphisms} for vector fields, the structural constraints imposed on the lumped Jacobian $\mathbb{J}_{\bm{\uptheta}}^{[i]}$ and the composition function $f{\mathbb{NN}}_\mathrm{dl}^{[i]}$ represent two complementary expressions of the same underlying algebraic property. Global convergence therefore dictates that every constituent sub-mapping
\[
  f{\mathbb{NN}}_k, \quad k \in \{i, i+1, \dots, n\},
\]
preserves strict monotonicity. In a standard artificial neural network, strict monotonicity of the activation functions ensures that their diagonal derivative factors do not change sign. It does not, however, determine the sign of the full lumped Jacobian, which also contains the layer-weight matrices. Condition \textbf{C1} must therefore be imposed independently of activation monotonicity.

Because the analytical deduction presented above is isomorphic for any arbitrary layer $i$, global system stability is fundamentally predicated on establishing a global diffeomorphism across all sequential layer operators:
\begin{equation}
    f{\mathbb{NN}}_i, \quad i \in \{1, 2, \dots, n\}
 \label{eqn_GlStb_Cndt}
\end{equation}
The fulfillment of this topological requirement therefore requires both strict monotonicity of the activation functions and the independent non-vanishing, sign-definite Jacobian condition, culminating in the following global convergence theorem.

\begin{theorem*}[\textbf{Global Cascaded Convergence and Uniqueness}] \label{thm_global_convergence}
Consider a deep cascaded neural network online learning system governed by the tracking energy function $L(\bm{\uptheta}) = \frac{1}{2}\|\mathbf{y} - f{\mathbb{NN}}_\mathrm{dl}(\mathbf{x}, \bm{\uptheta})\|^2$. Assume that all hidden-layer activation functions are strictly monotonic and that each associated lumped Jacobian matrix $\mathbb{J}_{\bm{\uptheta}}^{[i]}$, $i \in \{1, 2, \dots, n\}$, remains non-vanishing and strictly sign-definite across the entire operating domain. Under these topological conditions, the real-time system trajectories converge globally to a unique, time-varying optimal parameter trajectory $\bm{\uptheta}^*(t)$.
\end{theorem*}

\noindent \textbf{Parameter Learning Parallel Stability.} The global cascaded convergence and uniqueness theorem shows that, under active damping, monotonic activation functions, and the Jacobian conditions stated above, the parameter error $\delta\bm{\uptheta}^{[i]}$ converges to zero. This convergence constrains the high-dimensional manifold in Eqn\eqref{eqn_ey_TaylorSS} to the lower-dimensional subspace characterized by $\delta\bm{\uptheta}^{[i]} = \mathbf{0}$. However, as defined by Eqn\eqref{eqn_Theta_Matrix}, the parameter $\bm{\uptheta}^{[i]}$ is a local composition of vector space spanned by $\bm{\uptheta}_k^{[i]}$ for $k \in \{1, 2, \dots, n_i\}$. Consequently, the manifold Eqn\eqref{eqn_ey_TaylorSS} is reduced to
\begin{equation}
    {\mathbb{J}_{\uptheta_1}^{[i]}}^T \delta\bm{\uptheta}_1^{[i]} + {\mathbb{J}_{\uptheta_2}^{[i]}}^T \delta\bm{\uptheta}_2^{[i]} + \cdots + {\mathbb{J}_{\uptheta_k}^{[i]}}^T \delta\bm{\uptheta}_k^{[i]}+ \cdots + {\mathbb{J}_{\uptheta_{n_i}}^{[i]}}^T \delta\bm{\uptheta}_{n_i}^{[i]} = \textbf{0}
\label{eqn_manifold_Layer}
\end{equation}

\begin{figure}[h]
	\centering
	\includegraphics[scale=0.85]{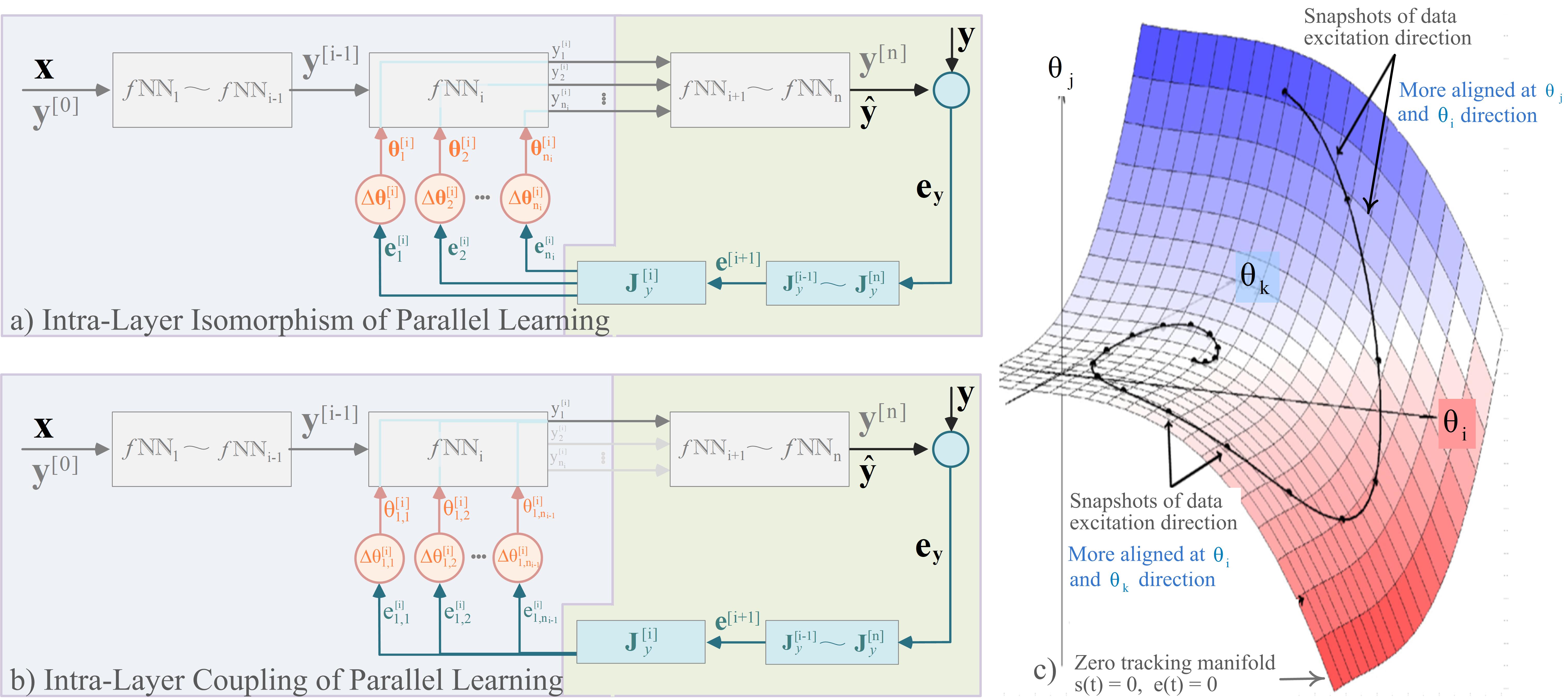}
	\caption{Topological Isomorphism of Intra-Layer Feedback Learning}
	\label{fig:IsoLayer}
\end{figure}

As illustrated in Fig.\ref{fig:IsoLayer}a, this represents a parallel learning structure wherein the individual learning loops are locally independent and isomorphic. By treating the downstream inference and error-construction layers as the physical world seen by the current layer, the independence of the parallel learning becomes apparent. Under the active damping, the convergence and stability of these parallel learning loops are guaranteed, provided the associated activation functions remain monotonic. To formalize this stability, the total energy function of Eqn\eqref{eqn_Energy_OverAll} is expanded as
\begin{equation}
    \mathcal{L}(\mathbf{y},\bm{\uptheta},t) = \frac{1}{2}\mathbf{e}_{y}^{T}\mathbf{e}_{y} + \sum_{i=1}^{n}\frac{1}{2\lambda_{i}}\sum_{k=1}^{n_i} {\delta \bm{\uptheta}_k^{[i]}}^{T}\delta \bm{\uptheta}_k^{[i]}
    \label{eqn_Energy_Expand}
\end{equation}
Here, the parallel learning components within a given layer are assumed to share a uniform learning rate, denoted by $\lambda_i$. Notably, this uniformity is not a strict requirement. Distinct localized learning rates $\lambda_{i,k}$ can be implemented without loss of generality. Evaluating the time derivative of the energy function via the same mathematical formalism established for cascaded learning yields
\begin{equation}
  \dot{\mathcal{L}}(t) = -\mathbf{e}_{y}^{T}\mathbb{K}_{y}\mathbf{e}_{y} - \sum_{i=1}^{n}\frac{1}{\lambda_{i}} \sum_{k=1}^{n_i} {\delta\bm{\uptheta}_k^{[i]}}^{T}\mathbb{K}_{\uptheta_k}^{[i]} \delta \bm{\uptheta}_k^{[i]} \leq 0
  \label{eqn_GlExp_SBT}
\end{equation}
According to Lyapunov stability criteria, Eqn\eqref{eqn_GlExp_SBT} implies that  $\delta\bm{\uptheta}_k^{[i]} \rightarrow \mathbf{0}$ asymptotically for all $k \in \{1, 2, \dots, n_i\}$.

Extending this framework to intra-layer parallel learning further confines the parameter space to the manifold defined by
\begin{equation}
    {\mathbb{J}_{\uptheta_{k,1}}^{[i]}} \delta\uptheta_{k,1}^{[i]} +
    {\mathbb{J}_{\uptheta_{k,2}}^{[i]}} \delta\uptheta_{k,2}^{[i]} + \cdots + {\mathbb{J}_{\uptheta_{k,l}}^{[i]}} \delta\uptheta_{k,l}^{[i]}+ \cdots + {\mathbb{J}_{\uptheta_{k,n_{i-1}}}^{[i]}} \delta\uptheta_{k,n_{i-1}}^{[i]} = 0
\label{eqn_manifold_Node}
\end{equation}
which represents the localized counterpart to the layer-level manifold in Eqn\eqref{eqn_manifold_Layer}. Because the variables in the above equation are scalars, explicit vector-transpose notation is unnecessary. This formulation establishes parameter convergence at the node level. From a system-dynamics perspective, node-level learning stability can be formalized through an approach isomorphic to the layer-level derivation. To account for these localized parameter estimation errors, the Lyapunov energy candidate of Eqn\eqref{eqn_Energy_Expand} is expanded as
\begin{equation}
    \mathcal{L}(\mathbf{y},\bm{\uptheta},t) = \frac{1}{2}\mathbf{e}_{y}^{T}\mathbf{e}_{y} + \sum_{i=1}^{n}\frac{1}{2\lambda_{i}}\sum_{k=1}^{n_i} \sum_{l=1}^{n_{i-1}}{\delta {{\uptheta}_{k,l}^{[i]}}}^2
    \label{eqn_Energy_Node}
\end{equation}
Evaluating the time derivative of this expanded function along the system trajectories yields
\begin{equation}
  \dot{\mathcal{L}}(t) = -\mathbf{e}_{y}^{T}\mathbb{K}_{y}\mathbf{e}_{y} - \sum_{i=1}^{n}\frac{1}{\lambda_{i}} \sum_{k=1}^{n_i} \sum_{l=1}^{n_{i-1}} \mathbb{K}_{\uptheta_{k,l}}^{[i]} {\delta {\uptheta}_{k,l}^{[i]}}^2 \leq 0
  \label{eqn_GlExpNode_SBT}
\end{equation}
Once again, under Lyapunov stability criteria, Eqn\eqref{eqn_GlExpNode_SBT} guarantees that the localized parameter deviations vanish asymptotically, implying that  $\delta{\uptheta}_{k,l}^{[i]} \rightarrow 0$ asymptotically for all $k \in \{1, 2, \dots, n_i\}$ and $l \in \{1, 2, \dots, n_{i-1}\}$.

Cross-referencing the error-dynamic manifolds given by  Eqn\eqref{eqn_ey_TaylorSS} with Eqn\eqref{eqn_manifold_Layer} and Eqn\eqref{eqn_manifold_Node}, as well as energy functionals of Eqn\eqref{eqn_Energy_OverAll} with Eqn\eqref{eqn_Energy_Expand} and Eqn\eqref{eqn_Energy_Node}, it reveals a fundamental mathematical structural isomorphism of parameter learning across the node, layer, and network levels. Node and layer level learning dynamics correspond structurally to the scalar and vector expansions of the block elements $\mathbb{K}_\uptheta^{[i]}$ and $\mathbb{J}_{\uptheta}^{[i]}$ in the state-space matrix $\mathbf{A}$ of Eqn\eqref{eqn_sysDyn_new}. Intrinsically, the invariant block structure of $\mathbf{A}$ governs global asymptotic stability.

However, while this mathematical proof confirms global stability, it masks the underlying coupled nature of node-level learning. Unlike the layer-level architecture where individual loops operate independently, learning at the node level is inherently coupled. As illustrated in Fig.~\ref{fig:IsoLayer}b, this multivariable interdependence is driven simultaneously by both forward information cascades and backward error-propagation pathways.

To reveal the data-driven nature of learning at the node level, the equivalent local learning system is abstracted as follows, in analogy with the layer-level derivation.
\begin{equation}
   \hat{y}(t)=f[\bm{\theta}^{T}(t)\mathbf{x}(t)]
    \label{eqn_MdlP_node}
\end{equation}
where $\mathbf{x}(t)$ is the inputs to the localized learning system at the node-level, and $\bm{\theta}$ the learning parameters as a generalized representation of ${\uptheta}_{k,l}^{[i]}$ locally. The output of $\hat{y}(t)$ shall follow the ideal $y(t)$ as
\[
   {y}(t)=f[{\bm\theta}^{*T}(t)\mathbf{x}(t)]
\]
in which $\bm{\theta}^*(t)$ is for $\bm{\theta}(t)$ to learn. $f: \mathbb{R} \to \mathbb{R}$ is a strictly monotonic and globally Lipschitz continuous activation function of the node, satisfying
\[
LP_{\max }\ge f'(\chi) \ge LP_{\min }>0
\]
Then, non-linear prediction error is as
\begin{equation}
   e(t)=y(t)-\hat{y}(t)
       =f[{\bm\theta}^{*T}(t)\mathbf{x}(t)]-f[\bm{\theta}^{T}(t)\mathbf{x}(t)]
       = \frac{\partial \hat{y}}{\partial \bm{\theta}}^{T} \delta{\bm\theta}(t)
 \label{eqn_error_node}
\end{equation}
The parameter error is defined as $\delta{\bm\theta}(t)= {\bm\theta}^{*}(t) - {\bm\theta}(t)$. Defining the energy function as $\mathcal{L}=\frac{1}{2} e^{2}(t)$, the minimum-energy principle gives the localized learning law
\begin{equation}
   \dot{\bm{\theta}}(t)= - \lambda \frac{\partial \mathcal{L}}{\partial \bm{\theta}}
   = \lambda e(t) \frac{\partial \hat{y}}{\partial \bm{\theta}}
   = \lambda \frac{\partial \hat{y}}{\partial \bm{\theta}} \frac{\partial \hat{y}}{\partial \bm{\theta}}^{T} \delta{\bm\theta}(t)
   = \lambda [f'(t)]^2 \mathbf{x}(t) \mathbf{x}^T(t) \delta{\bm\theta}(t)
   \label{eqn_learning_node}
\end{equation}
This symmetric matrix $\mathbf{x}(t) \mathbf{x}^T(t)$ acts as a directional filter for learning. For stability analysis, the positive-definite Lyapunov candidate is
\[
V(\delta{\bm\theta}) = \frac{1}{2\lambda}\delta{\bm\theta}^{T}(t)\delta{\bm\theta}(t)
\]
Bearing in mind that $\delta\dot{\bm\theta} (t) = -\dot{\bm{\theta}}(t)$, the time derivative along the non-linear trajectory becomes
\[
\dot{V} = \frac{1}{\lambda} \delta{\bm{\theta}}^{T}(t) \delta\dot{\bm{\theta}}(t)
= -[f'(t)]^2 \delta\bm{\theta}^{T}(t) \mathbf{x}(t) \mathbf{x}^T(t) \delta\bm{\theta}(t)
\]
Since $f'(t) \geq LP_{\min} > 0$ and the remaining quadratic term is non-negative, we obtain
\begin{equation}
   \dot{V}\le -LP_{\min }^2 \parallel \mathbf{x}^T(t) \delta\bm{\theta}(t) \parallel_2^2
   \;\le 0
   \label{eqn_learningSTB_node}
\end{equation}
This guarantees Lyapunov stability, $\delta\bm{\theta}(t)$ is bounded, and the asymptotical solution is
\begin{equation}
   \mathbf{x}^{T}(t) \delta\bm{\theta}(t) = 0
   \label{eqn_NullSpace_node}
\end{equation}
Since $e(t) = \frac{\partial \hat{y}}{\partial \bm{\theta}}^{T} \delta\bm{\theta}(t)$, and $\dot{e}(t) = \frac{\partial \hat{y}}{\partial \bm{\theta}}^{T} \delta\dot{\bm{\theta}}(t)$, then
\[
\dot{\mathcal{L}}= - e(t) \frac{\partial \hat{y}}{\partial \bm{\theta}}^{T} \frac{\partial \hat{y}}{\partial \bm{\theta}}e(t) \le 0
\]
By Barbalat's Lemma
\[
\lim _{t\rightarrow \infty }e(t)=0
\]
While the system effectively minimizes the energy functional $\mathcal{L}$, convergence of $\delta\bm{\theta}(t) = \mathbf{0}$ is not guaranteed. In particular, even when the tracking error vanishes ($e \to 0$), parameter adaptation can stall. Recall that the continuous-time parameter error dynamics evolve as a time-varying linear system governed by
\[
\delta\dot{\bm{\theta}}(t)=-\lambda [f'(t)]^2 \mathbf{x}(t)\mathbf{x}^{T}(t) \delta \bm{\theta }(t)
\]
The matrix $\bm\Omega(t) =\mathbf{x}(t)\mathbf{x}^{T}(t)$ is an outer product of an $l \times 1$ vector. For any given instant $t$, its rank is at most unity. Consequently, for an $l$-parameter system ($l > 1$), this matrix possesses a high-dimensional null space of dimension $l-1$. If the parameter error vector $\delta\bm{\theta}(t)$ projects into the null space of $\bm\Omega(t)$, the matrix operation collapses, yielding
\[
\delta\dot{\bm{\theta}}(t)=-\lambda \mathbf{x}(t) \underbrace{\left[\mathbf{x}^{T}(t)\delta \bm{\theta}(t)\right]}_{e(t)=0}=\mathbf{0}
\]
Because $\delta\dot{\bm{\theta}}(t) = \mathbf{0}$, the adaptation process completely freezes. The system cannot resolve any parameter errors orthogonal to the instantaneous input vector $\mathbf{x}(t)$, causing the algorithm to stagnate at an incorrect parameter estimate. Geometrically, the learning process can be interpreted as a trajectory migrating through an $l$-dimensional parameter error manifold. When the input vector $\mathbf{x}(t)$ lacks sufficient variation, the landscape of the energy function $\mathcal{L}$ forms a flat-bottomed valley consisting of an affine $l-1$ subspace. On this sub-manifold, the tracking error is identically zero, $e(t) = 0$, yet the parameter error remains non-zero, $\delta \bm{\theta }(t) \neq \mathbf{0}$. The error trajectory quickly descends the steep walls of this valley. However, upon intercepting the floor, tracking error feedback is completely eliminated, permanently trapping the parameter error within this unidentifiable subspace.

To break this null-space trap, the input signal $\mathbf{x}(t)$ must be Persistently Exciting (PE). A vector signal $\mathbf{x}(t)$ is PE if there exist constants $\alpha, T > 0$ such that
\begin{equation}
   \bm\Omega_T(t) = \int_{t}^{t+T}\mathbf{x}(\tau)\mathbf{x}^{T}(\tau) d\tau \ge \alpha I
   \label{eqn_PECndt_node}
\end{equation}
for all $t \geq 0$.
This requires the resulting matrix $\bm\Omega_T(t)$ to be full rank and strictly positive definite ($\ge \alpha I$) when the input data are integrated over a moving time window $T$. The input signal must change direction sufficiently to span the entire parameter space. To evaluate how the PE condition resolves the bottleneck for parameter learning convergence, the energy dissipation rate $\dot{V}(t)$ is integrated over a finite temporal window $[t, t+T]$
\[
V(t+T) - V(t) = \int_{t}^{t+T} \dot{V}(\tau) d\tau = -\int_{t}^{t+T} \left[f'(\tau)\right]^2 \delta\bm{\theta}^{T}(\tau) \bm\Omega(\tau) \delta\bm{\theta}(\tau) d\tau
\]

Under the assumption of a sufficiently small adaptation gain $\lambda$, the parameter error vector evolves slowly relative to the window duration $T$, permitting the approximation $\delta\bm{\theta}(\tau) \approx \delta\bm{\theta}(t)$ over the interval. Furthermore, the global diffeomorphism requires that the activation derivative remains bounded away from zero, such that $\left[f'(\tau)\right]^2 \geq \beta > 0$. This separation of timescales allows the parameter error vector to be factored out of the integral operator
\[
V(t+T) - V(t) \approx -\beta \delta\bm{\theta}^T(t) \left[ \int_{t}^{t+T} \bm\Omega(\tau) d\tau \right] \delta\bm{\theta}(t) = -\beta \delta\bm{\theta}^T(t) \bm\Omega_T(t) \delta\bm{\theta}(t)
\]

Substituting the lower bound established by the PE condition, $\bm\Omega_T(t) \geq \alpha \mathbf{I}$, yields the following contractive inequality:
\[
V(t+T) - V(t) \leq -\alpha \beta \delta\bm{\theta}^T(t) \delta\bm{\theta}(t)
\]

To express this relationship in a standard continuous-time Lyapunov format, the finite-time temporal expectation operator $\mathbb{E}_T[\cdot]$ is applied, where $\mathbb{E}_T[\dot{V}(t)] = \frac{1}{T}\left(V(t+T) - V(t)\right)$, translating the inequality into
\begin{equation}
   \mathbb{E}_T[\dot{V}(t)] \leq -\frac{\alpha \beta}{T} \delta\bm{\theta}^T(t) \delta\bm{\theta}(t)
   \label{eqn_PEStb_node}
\end{equation}

Because $V(t)$ is lower-bounded by zero and $\mathbb{E}_T[\dot{V}(t)]$ is strictly negative semi-definite, the integral of the mean dissipation rate remains finite. According to Barbalat's Lemma, the uniform continuity of the system trajectories ensures that this average dissipation rate asymptotically vanishes. Consequently, the parameter learning error is guaranteed to converge asymptotically to the origin, ensuring exact parameter identification
\[
\lim_{t \to \infty} \mathbb{E}_T[\dot{V}(t)] = 0 \implies \lim_{t \to \infty} \delta\bm{\theta}(t) = \mathbf{0} \implies \lim_{t \to \infty} \bm{\theta}(t) = \bm{\theta}^*
\]

This adaptive learning convergence exhibits a close analogy to sliding-mode systems (SMS) \cite{utkin2017sliding}, where trajectories are constrained to a predefined manifold. Driven by large initial tracking errors, the instantaneous gradient generates an orthogonal restoring force that rapidly forces the parameter error vector onto the time-varying hyper-surface
\begin{equation}
   s(t) = \delta\bm{\theta}^T(t)\mathbf{x}(t) = 0
   \label{eqn_errManifold_node}
\end{equation}
which is the same as $e(t) = 0$ according to Eqn\eqref{eqn_error_node}. However, a fundamental divergence from traditional sliding-mode systems occurs upon interception. Whereas conventional sliding-mode trajectories navigate toward the origin via autonomous internal dynamics, error-feedback gradients vanish identically on $s(t) = 0$, rendering the trajectory instantly stagnant.

PE breaks this stasis by modulating the kernel of $s(t) = 0$ continuously. As the regressor vector $\mathbf{x}(t)$ varies continuously under the PE condition, this kernel undergoes a dynamic rotation. Any directional perturbation immediately displaces the parameter state out of the manifold, generating a non-zero tracking error [$e(t) \neq 0$] that reactivates the gradient field and produces a localized contractive force. Topologically, the unique invariant intersection of all these dynamically rotating hyper-planes over any non-vanishing interval contracts exclusively to the origin $\delta\bm{\theta}(t) = \mathbf{0}$. This sliding-mode-like mechanism inherent to data-driven learning is illustrated as Fig.\ref{fig:IsoLayer}c, where continuous directional excitation prevents parameter stasis by constraining the trajectory to the zero-tracking manifold.

\begin{theorem*}[\textbf{Continuous Online Learning Global Stability}]
\label{thm_global_stability}
Consider a deep layered neural network of $\mathbf{\hat{y}} =f{\mathbb{NN}}_\mathrm{dl}(\mathbf{x}, \bm{\uptheta})$, where the parameter adaptation is governed by the tracking error feedback of $\dot{\bm{\uptheta}}^{[i]}+ \mathbb{K}_\uptheta^{[i]} \bm{\uptheta}^{[i]} = \lambda_i \mathbb{J}_\uptheta^{[i]} \mathbf{e}_y$. If all hidden-layer activation functions are strictly monotonic, each lumped Jacobian $\mathbb{J}_{\bm{\uptheta}}^{[i]}$, $i \in \{1,2,\dots,n\}$, remains non-vanishing and strictly sign-definite across the entire operating domain, and the regressor is persistently exciting in the sense of Eqn~(\ref{eqn_PECndt_node}), the online learning dynamics are asymptotically stable under the Lyapunov energy principle.
\end{theorem*}

\section{Conclusion}

\hspace*{\parindent}This paper analyses Online Active Learning (\textbf{OAL}) as a stability problem for large-scale neural systems. A decreasing prediction error alone does not establish stable online learning. Multilayer networks admit high-dimensional manifolds of parameter configurations that produce identical outputs. Parameter convergence, cross-layer coupling, and internal zero dynamics must therefore be analysed explicitly.

Four analytical results are established. First, under bounded external variation, the \textbf{OAL} error-feedback law makes the prediction-error energy non-increasing, as shown in Eqn(\ref{eqn_OAL_GAStb}). Second, the block state-space representation in Eqn(\ref{eqn_sysDyn}) shows no direct parameter-to-parameter coupling between different layers. Their interaction is mediated by the shared error $\mathbf{e}_y$. These cross-layer terms are skew-symmetric and therefore energy-conservative. Spectral analysis shows that the outer tracking loop is damped, whereas the internal zero dynamics remain marginally stable in $\ker(\mathds{J}_{sk})$.

Third, the analysis identifies persistency of excitation and active damping as two mechanisms for removing this marginal stability. Under active damping, the cascaded Lyapunov analysis in Eqn(\ref{eqn_GlLyp_SBT}) gives $\mathbf{e}_y \rightarrow \mathbf{0}$ and $\delta\bm{\uptheta}^{[i]} \rightarrow \mathbf{0}$ for every layer. The isomorphic argument extends to intra-layer and node-level parameters, where the \textbf{PE} condition contracts the zero-tracking manifold onto the origin. Fourth, uniqueness of the time-varying target trajectory $\bm{\uptheta}^*(t)$ requires diffeomorphic layer mappings. For the neural architecture considered here, this requires strictly monotonic activation functions together with non-vanishing, sign-definite lumped Jacobians. In the simplified damping case, the parameter energy decays as $e^{-2\kappa t}$, so $\kappa$ sets the exponential decay rate.

These results are analytical and require experimental validation. Future work will run the primary framework of Fig.~\ref{fig:OALPrimary}b on continuously arriving sensory streams. The experiments will measure prediction-error decay and, where an identifiable reference is available, parameter-deviation decay during uninterrupted online operation. They will also test whether natural streams satisfy Eqn(\ref{eqn_PE_Condition}) and quantify how $\kappa$ balances retention of learned structure against adaptation speed. Comparisons with offline-trained models of the same capacity will assess whether monotonic activations, sufficient excitation, and appropriate damping are achieved in practical vision-based perception systems for autonomous driving and robotics.
\bibliography{references}

\end{document}